\documentclass[sigconf]{acmart}

\usepackage[utf8]{inputenc} 
\usepackage[T1]{fontenc}    
\usepackage{booktabs}       
\usepackage{amsfonts}       
\usepackage{nicefrac}       
\usepackage{microtype}      
\usepackage{amsmath,amsfonts,amsthm,commath}
\usepackage{enumerate}
\usepackage{framed}
\usepackage{xspace}
\usepackage{microtype}
\usepackage{nicefrac}
\usepackage[utf8]{inputenc}
\usepackage{CJK}
\usepackage{indentfirst}
\usepackage{booktabs,color} 
\usepackage{fancyhdr}
\usepackage{graphicx}
\usepackage[tight,footnotesize]{subfigure}
\usepackage{multirow}
\usepackage{listings}
\usepackage{color}
\usepackage{caption}
\usepackage{algorithm}
\usepackage{algorithmicx}
\usepackage[noend]{algpseudocode}
\usepackage{xspace}
\usepackage{url}
\usepackage{breakurl}

\usepackage{flushend}
\usepackage{xspace,color}
\usepackage{bm}

\newcommand{\sysn}{MuFasa\xspace}

\newcommand{\bx}{\mathbf{x}}
\newcommand{\bxt}{\mathbf{x}_t}
\newcommand{\bxtk}{\mathbf{x}_t^k}

\newcommand{\arm}{\mathbf{x}^k_{t, i}}
\newcommand{\arms}{\mathbf{X}^k_t}
\newcommand{\sarm}{\mathbf{x}^k_t}
\newcommand{\eps}{\epsilon_t}
\newcommand{\rt}{\mathbf{r}_t}

\newcommand{\armc}{\mathbf{X}_t}
\newcommand{\ksarm}{\mathbf{X}_t}
\newcommand{\bthetas}{\boldsymbol{\theta}^{\Sigma}}
\newcommand{\ve}{\text{vec}}

\newcommand{\st}{\mathbf{S}_t}
\newcommand{\btheta}{\boldsymbol{\theta}}

\newcommand{\bsigma}{\boldsymbol{\Sigma}}
\newcommand{\trains}{\{\mathbf{x}_i\}_{i=1}^{T}}
\newcommand{\bts}{\boldsymbol{\theta}^{\ast}}

\newcommand{\gx}{g(\mathbf{x}_t; \btheta_0)}

\newcommand{\deter}{\text{\normalfont det}}

\newcommand{\fx}{f(\mathbf{x}_t; \btheta_t)}
\newcommand{\hx}{h(\mathbf{x}_t)}

\newtheorem{prove}{Prove}[section]

\newcommand{\para}[1]{\xspace \smallskip \noindent\textbf{#1}\xspace}


\AtBeginDocument{%
  \providecommand\BibTeX{{%
    \normalfont B\kern-0.5em{\scshape i\kern-0.25em b}\kern-0.8em\TeX}}}

\copyrightyear{2021} 
\acmYear{2021} 
\setcopyright{acmcopyright}\acmConference[KDD '21]{Proceedings of the 27th ACM SIGKDD Conference on Knowledge Discovery and Data Mining}{August 14--18, 2021}{Virtual Event, Singapore}
\acmBooktitle{Proceedings of the 27th ACM SIGKDD Conference on Knowledge Discovery and Data Mining (KDD '21), August 14--18, 2021, Virtual Event, Singapore}
\acmPrice{15.00}
\acmDOI{10.1145/3447548.3467299}
\acmISBN{978-1-4503-8332-5/21/08}

\begin{document}
\fancyhead{}
\title{Multi-facet Contextual Bandits: A Neural Network Perspective}



\author{Yikun Ban}
\affiliation{%
  \institution{University of Illinois at Urbana-Champaign}
  \country{}
  }
\email{yikunb2@illinois.edu}

\author{Jingrui He}
\affiliation{%
  \institution{University of Illinois at Urbana-Champaign}
  \country{}
}
\email{jingrui@illinois.edu  }

\author{Curtiss B. Cook}
\affiliation{%
  \institution{Mayo Clinic Arizona}
  \country{}
}
\email{cook.curtiss@mayo.edu}

\begin{abstract}

Contextual multi-armed bandit has shown to be an effective tool in recommender systems. In this paper, we study a novel problem of multi-facet bandits involving a group of bandits, each characterizing the users' needs from one unique aspect. In each round, for the given user, we need to select one arm from each bandit, such that the combination of all arms maximizes the final reward. This problem can find immediate applications in E-commerce, healthcare, etc. To address this problem, we propose a novel algorithm, named \sysn, which utilizes an assembled neural network to jointly learn the underlying reward functions of multiple bandits. It estimates an Upper Confidence Bound (UCB) linked with the expected reward to balance between exploitation and exploration. Under mild assumptions, we provide the regret analysis of \sysn. It can achieve the near-optimal $\widetilde{ \mathcal{O}}((K+1)\sqrt{T})$ regret bound where $K$ is the number of bandits and $T$ is the number of played rounds. Furthermore, we conduct extensive experiments to show that \sysn outperforms strong baselines on real-world data sets.

\end{abstract}




\maketitle

\section{introduction}
The personalized recommendation is ubiquitous in web applications. 
Conventional approaches that rely on sufficient historical records, e.g., collaborative filtering \cite{su2009survey, zhou2020gan}, have proven successful both theoretically and empirically. 
However, with the cold-start problem and the rapid change of the recommendation content, these methods might render sub-optimal performance~\cite{2010contextual, 2014onlinecluster}. To solve the dilemma between the exploitation of historical data and the exploration of new information, Multi-Armed Bandit (MAB) \cite{langford2008epoch,auer2002finite,2011improved,auer2002nonstochastic,ban2020generic}  turns out to be an effective tool, which has been adapted to personalized recommendation \cite{2010contextual, 2014onlinecluster}, online advertising~\cite{wu2016contextual}, clinical trials \cite{2018durandcontextual, 2020bastanionline}, etc.

In the conventional contextual bandit problem setting \cite{2010contextual}, i.e., single MAB, the learner is presented with a set of arms in each round, where each arm is represented by a feature vector.
Then the learner needs to select and play one arm to receive the corresponding reward that is drawn from an unknown distribution with an unknown mean.
To achieve the goal of maximizing the accumulated rewards, the learner needs to consider the arms with the best historical feedback as well as the new arms for potential gains. 
The single MAB problem has been well studied in various settings.
With respect to the reward function, one research direction  \cite{2010contextual, 2011improved,dimakopoulou2019balanced, 2014onlinecluster,2016collaborative} assumes that the expected reward is linear with respect to the arm's feature vector. 
However, in many real applications, this assumption fails to hold. 
Thus many exiting works turn to focus on the nonlinear or nonparametric bandits \cite{srinivas2009gaussian, bubeck2011x} with mild assumptions such as the Lipschitz continuous property \cite{bubeck2011x} or embedding in Reproducing Kernel Hilbert Space \cite{valko2013finite, deshmukh2017multi}.
Furthermore, the single MAB problem has been extended to best arm identification ~\cite{auer2002finite,2010best}, outlier arm identification~\cite{gentile2017context, ban2020generic}, Top-K arm problems ~\cite{buccapatnam2013multi}, and so on.

In this paper, we define and study a novel problem of \textit{multi-facet contextual bandits}. In this problem, the users' needs are characterized from multiple aspects, each associated with one bandit.
Consider a task consisting of $K$ bandits, where each bandit presents a set of arms separately and the learner needs to choose and play one arm from each bandit. 
Therefore, a total of $K$ arms are played in one round. 
In accordance with the standard bandit problem, the learner can observe a reward after playing one arm from the bandit, which we call "sub-reward", and thus $K$ sub-rewards are received in total. 
In addition, a reward that is a function with respect to these $K$ sub-rewards, called "final reward", is observed to represent the overall feedback with respect to the $K$ selected arms.
Note that the functions of final reward and $K$ sub-rewards are allowed to be either linear or non-linear.
The goal of the learner in the multi-facet bandit problem is to maximize the final rewards of all the played rounds.

This problem finds many applications in real-world problems.
For instance, in the recommender system, instead of the single item recommendation, an E-commerce company launches a promotion campaign, which sells collections of multiple types of products such as snacks, toiletries, and beverages.
Each type of item can be formulated as a multi-armed bandit and the learner aims to select the best combination of snack, toiletry, and beverage. 
As a result, the final reward is the review of this combined recommendation, while the sub-reward is the review for a particular product
. This problem also exists in healthcare. For a diabetes patient, the doctor usually provides a comprehensive recommendation including medication, daily diet, and exercise, where each type has several options. 
Here, the final reward can be set as the change of key biomarkers for diabetes (e.g., HbA1c) and the sub-reward can be the direct impact of each type of recommendation (e.g., blood pressure change for a medicine).

A major challenge of the proposed multi-facet bandit problem is the partial availability of sub-rewards, as not every sub-reward is easy to observe.
For example, regarding the combined recommendation of E-commerce, the user may rate the combination but not individual items; regarding the comprehensive recommendation for a diabetes patient, some sub-rewards can be difficult to measure (e.g., the impact of low-calorie diets on the patient's overall health conditions). Therefore, in our work, we allow only a subset of all sub-rewards to be observed in each round, which increases the flexibility of our proposed framework.

To address these challenges, we aim to learn the mappings from the selected $K$ arms (one from each bandit) to the final rewards, incorporating two crucial factors: (1) the collaborative relations exist among these bandits as they formulate the aspects from one same user; (2) the bandits contribute to the task with various weights because some aspects (bandits) are decisive while some maybe not. 
Hence, we propose a novel algorithm, \sysn, to learn $K$ bandits jointly. It utilizes an assembled neural networks to learn the final reward function combined with $K$ bandits.  
Although the neural networks have been adapted to the bandit problem \cite{riquelme2018deep, zahavy2019deep,zhou2020neural}, they are designed for the single bandit with one selected arm and one reward in each round. 
To balance the exploitation and exploration of arm sets, we provide a comprehensive upper confidence bound based on the assembled network linking the predicted reward with the expected reward. When the sub-rewards are partially available, we introduce a new approach to leverage them to train bandits jointly. Furthermore, we carry out the theoretical analysis of \sysn and prove a near-optimal regret bound under mild assumptions.   
Our major contributions can be summarized as follows:

(1) \textbf{Problem}. We introduce the problem of multi-facet contextual bandits to characterize the users' needs from multiple aspects, which can find immediate applications in E-commerce, healthcare, etc.         
    
(2) \textbf{Algorithm}. We propose a novel algorithm, \sysn, which exploits the final reward and up to $K$ sub-rewards to train the assembled neural networks and explores potential arm sets with a UCB-based strategy.  
    
(3) \textbf{Theoretical analysis}. Under mild assumptions, we provide the upper confidence bounds for a neural network and the assembled neural networks. Then, we prove that \sysn can achieve the $\widetilde{\mathcal{O}}((K+1)\sqrt{T})$ regret bound, which is near-optimal compared to a single contextual bandit. 
    
(4) \textbf{Empirical performance}. We conduct extensive experiments to show the effectiveness of \sysn, which outperforms strong baselines on real-world data sets even with partial sub-rewards.

\section{related work}

\para{Multi-armed bandit.}
The multi-armed bandit was first introduced by ~\cite{thompson1933likelihood} and then further studied by many works that succeeded in both theory and practice such as $\epsilon$-greedy \cite{langford2008epoch}, Thompson sampling\cite{agrawal2013thompson}, and upper confidence bound \cite{auer2002finite}.
In the contrast with traditional bandits \cite{auer2002finite,ban2020generic}, the contextual bandit \cite{2010contextual, 2011improved,wu2016contextual} has the better representation capacity where each arm is represented by a context vector instead of a scalar to infer the reward. 
Among them, the linear contextual bandits are extensively studied and many of them use the UCB strategy, achieving $\widetilde{\mathcal{O}}(\sqrt{T})$ regret bound \cite{2011improved, 2014onlinecluster,ban2021local}. To further generalize the reward function, many works use a nonlinear regression model drawn from the reproducing kernel Hilbert space to learn the mapping from contexts to rewards such as the kernel-based methods \cite{deshmukh2017multi, valko2013finite}.

\para{Neural bandits.} 
The authors of~\cite{allesiardo2014neural} use a neural work to model an arm and then applied $\epsilon$-greedy strategy to select an arm. In contrast, \sysn utilizes a UCB-based strategy working on $K$ bandits instead of one set of arms. In addition, the Thompson sampling has been combined with deep neural networks \cite{lipton2018bbq,azizzadenesheli2018efficient,riquelme2018deep,zahavy2019deep}. For instance, \cite{riquelme2018deep,zahavy2019deep} regard the last layer of the neural network as the embeddings of contexts and then apply the Thompson sampling to play an arm in each round.  
NeuUCB \cite{zhou2020neural} first uses the UCB-based approach constructed on a fully-connected neural network, while it only fits on the single bandit with one set of arms. On the contrary, \sysn constructs an assembled neural networks to learn $K$ bandits jointly.
Deep neural network in multi-view learning has been well-studied \cite{zhou2015muvir,zhou2020domain,fu2020view, zheng2021deep,jing2021hdmi}, to extract useful information among multiple sources, which inspires one of the core ideas of \sysn.

\para{Other variant bandit setting.}
In the non-contextual bandit, a number of works \cite{chen2013combinatorial,chen2016combinatorial, liu2011logarithmic} study playing $K$ arms at the same time in a single bandit, while these approaches have limited representation power in the recommender system.
The most similar setting is the contextual combinatorial MAB problem\cite{qin2014contextual,li2016contextual}, where the learner tends to choose the optimal subset of arms with certain constraints like the $K$-size.
One key difference is that all the arms are from the same single bandit where only one reward function exists.
On the contrary, in the multi-faced bandits, the selected $K$ arms come from $K$ different bandits with $K$ different reward functions and the sub-rewards are allowed to be partially available.
There is another line of works \cite{2014onlinecluster, 2016collaborative, ban2021local} for bandit clustering, where a bandit is constructed for each user.  They try to leverage the dependency among users to improve the recommendation performance. 
However, in these works, they still play one arm in each round and the reward function is required to be linear.

\section{Problem Definition}
In this section, we formulate the problem of multi-facet bandits, with a total of $K$ bandits, where the learner aims to select the optimal set of $K$ arms in each round, in order to maximize the final accumulated rewards.

Suppose there are $T$ rounds altogether. In each round $t \in [T]$ ($[T] = \{1, \dots, T\}$), the learner is faced with $K$ bandits, and each bandit $k \in [K]$ has a set of arms $\arms =  \{\mathbf{x}^k_{t,1}, \dots, \mathbf{x}^{k}_{t, n_k} \}$, where $ |\arms| = n_k$ is the number of arms in this bandit.
In the bandit $k$, for each arm $\arm \in \arms$, it is represented by a $d_k$-dimensional feature vector and we assume $\| \arm \|_2 \leq 1$.
Subsequently, in each round $t$, the learner will observe $K$ arm sets $\{ \arms \}_{k=1}^K$ and thus a total of $\sum_{k=1}^K n_k$ arms. 
As only one arm can be played within each bandit, the learner needs to select and play $K$ arms denoted as $\ksarm = \{\mathbf{x}^1_t, \dots, \mathbf{x}^k_t,  \dots, \mathbf{x}^K_t\}$ in which $\bxtk \in \armc$  represents the selected arm from $\arms$.

Once the selected arm $\sarm$ is played for bandit $k$, a sub-reward  $r^k_t$ will be received to represent the feedback of this play for bandit $k$  separately.  The sub-reward is assumed to be governed by an unknown reward function:
\[
r^k_t(\sarm) = h_k(\sarm).
\]
where $h_k$ can be either a linear \cite{2010contextual, 2011improved} or non-linear reward  function \cite{deshmukh2017multi, valko2013finite}. As a result, in each round $t$, the learner needs to play $K$ arms in $\ksarm$ and then receive $K$ sub-rewards denoted by $\mathbf{r}_t = \{ r^1_t,  \dots, r^k_t, \dots, r^K_t \}$.

As the $K$ bandits characterize the users' needs from various aspects, after playing $K$ arms in each round $t$, a final reward $R_t$ will be received to represent the overall feedback of the group of $K$ bandits. 
The final reward $R_t$ is considered to be governed by an unknown function with respect to $\rt$:
\[
R_t(\rt) = H \left( \left( h_1(\bx_t^{1}), \dots,  h_k(\bx_t^{k})   \dots, h_K(\bx_t^{K}) \right ) \right) + \eps.
\]
where $\eps$ is a noise drawn from a Gaussian distribution with zero mean.  
In our analysis, we make the following assumptions regarding $h_k$ and $H(\ve(\mathbf{r}_t))$:
\begin{enumerate}
    \item If $\bx_t^k = \mathbf{0}$, then $h(\bx_t^k) = 0$;   If $\ve(\rt)  = (0, \dots, 0 )$, then $H( \ve(\rt)) = 0$.
    \item \textbf{$\bar{C}$-Lipschitz continuity}. $H(\ve(\rt))$ is assumed to be $\bar{C}$-Lipschitz continuous with respect to the $\mathbf{r}_t$. Formally, there exists a constant $\bar{C} > 0$ such that
    \[
    | H(\ve(\rt)) -   H(\ve(\mathbf{r}_t'))| \leq \bar{C} \sqrt{ \sum_{k \in K} [r_t^k - {r_t^k}']^2}.
    \]
\end{enumerate}
Both assumptions are mild. For (1), if the input is zero, then the reward should also be zero. For (2), the Lipschitz continuity can be applied to many real-world applications. For the convenience of presentation, given any set of selected $K$ arms $\armc$,  we denote the expectation of $R_t$ by:
\begin{equation} 
\mathcal{H}(\armc) = \mathbb{E}[R_t | \armc] = H \left( \left( h_1(\bx_t^{1}), \dots,  h_k(\bx_t^{k})   \dots, h_K(\bx_t^{K}) \right )  \right).
\end{equation}


Recall that in multi-facet bandits, the learner aims to select the optimal $K$ arms with the maximal final reward $ R_t^\ast$ in each round. First, we need to identify all possible combinations of $K$ arms, denoted by
\begin{equation} \label{eq:st}
\mathbf{S}_t = \{ (\mathbf{x}^1_t, \dots, \mathbf{x}^k_t,  \dots, \mathbf{x}^K_t) \ | \ \sarm \in \arms, k \in [K] \},
\end{equation}
where $|\st| = \prod_{k=1}^K n_k$ because bandit $k$ has  $n_k$ arms for each $k \in [K]$.
Thus, the regret of multi-facet bandit problem is defined as 
\[
\begin{aligned}
\textbf{Reg} & = \mathbb{E}[ \sum_{t=1}^T ( R_t^\ast - R_t)] \\
& = \sum_{t=1}^T  \left(   \mathcal{H}(\armc^{\ast}) - \mathcal{H}(\armc) \right),
\end{aligned}
\]
where $\armc^{\ast} = \arg \max_{  \mathbf{X}_t \in \st} \mathcal{H} (\mathbf{X}_t)$.
Therefore, our goal is to design a bandit algorithm to select $K$ arms every round in order to minimize the regret. We use the standard $\mathcal{O}$ to hide constants and $\widetilde{\mathcal{O}}$ to hide logarithm.

\para{Availability of sub-rewards}. 
In this framework, the final $R_t$ is required to be known, while the sub-rewards $\rt$ are allowed to be partially available. Because the feedback of some bandits cannot be directly measured or is simply not available in a real problem.  
This increases the flexibility of our proposed framework. 

More specifically, in each round $t$, ideally, the learner is able to receive $K + 1 $ rewards including $K$ sub-rewards $\{r^1_t, \dots, r^K_t\}$  and a final reward $R_t$. As the final reward is the integral feedback of the entire group of bandits and reflects how the bandits affect each other, $R_t$ is required to be known.  However, the $K$ sub-rewards are allowed to be partially available, because not every sub-reward is easy to obtain or can be measured accurately.  

This is a new challenge in the multi-facet bandit problem. Thus, to learn $\mathcal{H}$, the designed bandit algorithm is required to handle the partial availability of sub-rewards. 

\section{Proposed Algorithm}

In this section, we introduce the proposed algorithm, \sysn. The presentation of \sysn is divided into three parts. First, we present the neural network model used in \sysn; Second, we detail how to collect training samples to train the model in each round; In the end, we describe the UCB-based arm selection criterion and summarize the workflow of \sysn.   

\subsection{Neural network model}

To learn the reward function $\mathcal{H}$, we use $K+1$ fully-connected neural networks to learn $K$ bandits jointly, where a neural network $f_k$ is built for each bandit $k \in [K]$ to learn its reward function $h_k$, and a shared neural network $F$ is constructed to learn the mapping from the $K$ neural networks $(f_1, \dots, f_K)$ to the final reward $R_t$. 

First, in round $t$, for each bandit $k \in [K]$, given any context vector $\sarm \in \mathbb{R}^{d_k} $,  we use a $L_1$-layer fully-connected network to learn $h_k$ , denoted by $f_k$:
\[
f_k(\sarm;  \btheta^k) = \sqrt{m_1} \mathbf{W}_{L_1} \sigma (\mathbf{W}_{L_1 -1}\sigma(\dots \sigma(\mathbf{W}_1 \sarm))),
\]
where $\sigma(x)$ is the rectified linear unit (ReLU) activation function. Without loss of generality, we assume each layer has the same width $m_1$ for the sake of analysis. Therefore, $\btheta^k = \big ( \ve(\mathbf{W}_{L_1})^\intercal, \dots,$ \break $\ve(\mathbf{W}_1 )^\intercal \big )^\intercal$ $ \in \mathbb{R}^{P_1} $, where $\mathbf{W}_1 \in \mathbb{R}^{m_1 \times d_k}$,  $ \mathbf{W}_i \in \mathbb{R}^{m_1 \times m_1}, \forall i \in [1 : L_1-1]$, and $\mathbf{W}_{L_1} \in \mathbb{R}^{\widehat{m} \times m_1}$. 
Note that $f_k(\sarm;  \btheta^k) \in \mathbb{R}^{\widehat{m}}$, where $\widehat{m}$ is set as a tuneable parameter to connect with the following network $F$.  Denote the gradient $\triangledown_{\btheta^k} f_k(\sarm;  \btheta^k) $ by $g(\bxtk; \btheta^k) $. 

Next, to learn the final reward function $H$, we use a $L_2$-layer fully-connected network to combine the outputs of the above $K$ neural networks, denoted by $F$:
\[
F\left( \mathbf{f}_t ; \btheta^{\Sigma} \right) = \sqrt{m_2}  \mathbf{W}_{L_2} \sigma (\dots \sigma(\mathbf{W}_1 ( \mathbf{f}_t ) ))
\]
where  $ \mathbf{f}_t  = \left( f_1(\bx_t^1; \btheta^1 )^\intercal, \dots, f_K(\bx_t^K; \btheta^K )^\intercal \right)^\intercal \in \mathbb{R}^{\widehat{m}K}$. Also, we assume that each layer has the same width $m_2$. Therefore,  $\bthetas = ( \ve(\mathbf{W}_{L_2})^\intercal$ $, \dots, \ve(\mathbf{W}_1 )^\intercal )^\intercal \in \mathbb{R}^{P_2}$, where $\mathbf{W}_1 \in \mathbb{R}^{m_2 \times \widehat{m}K}$,  $ \mathbf{W}_i \in \mathbb{R}^{m_2 \times m_2}, \forall i \in [L_2-1]$ and  $\mathbf{W}_{L_2} \in \mathbb{R}^{1 \times m_2}$,
Denote the gradient $\triangledown_{\bthetas} F\left( \mathbf{f}_t ; \btheta^{\Sigma} \right)  $  by $G(\mathbf{f}_t; \bthetas)$.

Therefore, for the convenience of presentation,  the whole assembled neural networks can be represented by $\mathcal{F}$ to learn $\mathcal{H}$ (Eq.(\ref{eq:h12})), given the $K$ selected arms $\armc$:
\[
\mathcal{F}(\armc; \btheta) = \left( F( \cdot ;\btheta^{\Sigma}) \circ \left ( f_1( \cdot ; \btheta^1), \dots,    f_K( \cdot ; \btheta^K) \right ) \right ) (\armc), 
\]
where $\btheta = (\bthetas, \btheta^1, \dots, \btheta^K)$. 

\para{Initialization.} $\btheta$ is initialized by randomly generating each parameter from the Gaussian distribution. More specifically, for $\btheta^k, k \in [K]$, $ \mathbf{W}_l$ is set to $\begin{pmatrix}  
\mathbf{w} & \mathbf{0} \\ \mathbf{0}& \mathbf{w}
\end{pmatrix}$ for any $l \in [L_1]$ where $\mathbf{w}$ is drawn from $N(0, 4/m_1)$.
For $\bthetas$, $ \mathbf{W}_l$ is set to $\begin{pmatrix}  
\mathbf{w} & \mathbf{0} \\ \mathbf{0}& \mathbf{w}
\end{pmatrix}$ for any $l \in [L_2-1]$ where $\mathbf{w}$ is drawn from $N(0, 4/m_2)$;  $\mathbf{W}_{L_2} $ is set to $(\mathbf{w}^{\intercal}, -\mathbf{w}^{\intercal})$ where $\mathbf{w}$ is drawn from $N(0,2/m_2)$.

\newcommand{\gxk}{g(\bxtk; \btheta^k_t)}
\newcommand{\bgf}{G(\mathbf{f}_t; \bthetas_t)}

\begin{algorithm}[t]
\renewcommand{\algorithmicrequire}{\textbf{Input:}}
\renewcommand{\algorithmicensure}{\textbf{Output:}}
\caption{ \sysn }\label{alg:main}
\begin{algorithmic}[1] 
\Require $\mathcal{F}$,  $T$, $K$, $\delta$,  $\eta$ , $J$ 
\State  Initialize $\btheta_0 = (\bthetas_0, \btheta^1_0, \dots, \btheta^K_0)$  
\For{each $t \in [T]$}
\For{each bandit $k \in [K]$ }
\State Observe context vectors $\mathbf{X}_t^k = \{\mathbf{x}_{t,1}^k, \dots, \mathbf{x}_{t, n_k}^k \}$
\EndFor
\State Collect $\mathbf{S}_t$ (Eq. (\ref{eq:st}))
\State Choose $K$ arms, $\ksarm$, by:
 $$\ksarm = \arg \max_{\mathbf{X}_t' \in \mathbf{S}_t } \bigg (  \mathcal{F}( \mathbf{X}_t' ; \btheta_{t-1}) + \text{UCB}(\mathbf{X}_t')  \bigg  ).    \  \ \  ( \ \text{Theorem} \  \ref{theo:ucb} \ ) $$
\State  Play $\armc$ and observe rewards $R_t$ and  $ \mathbf{r}_t$.
\If{ $| \mathbf{r}_t| = K$ }  \#\#  \textit{  sub-rewards are all available. } 
\State  $\btheta_t$ = $GradientDescent_{All}$($\mathcal{F}$,$\{ \mathbf{X}_i \}_{i=1}^t$, $ \{R_i \}_{i=1}^t $, $ \{\mathbf{r}_i \}_{i=1}^t $ ,  $J, \eta$)
\Else \ \  \#\#  \textit{ sub-rewards are partially available. } 
\State Collect $\{ \Omega_i \}_{i=1}^t$ \  \  ( Eq.(\ref{eq:armpool}) )
\State $\btheta_t$ = $GradientDescent_{Partial}$ ($\mathcal{F}$,  $\{ \Omega_i \}_{i=1}^t$,  $J$, $\eta$)
\EndIf
\State Update $\text{UCB}(\mathbf{X}_t')$.
\EndFor
\end{algorithmic}
\end{algorithm}

\subsection{Training process}

Only with the final reward $R_t$ and $K$ selected arms $\armc$, the training of the neural network model $\mathcal{F}$ is the following minimization problem:
\begin{equation} \label{eq:lossf}
 \min_{\btheta} \mathcal{L} (\btheta) = \sum_{t=1}^T  \left(\mathcal{F}(\mathbf{X}_t; \btheta  )-    R_t\right)^2 /2 + m_2\lambda\|\btheta - \btheta_0\|_2^2 /2 .
\end{equation}
where $\mathcal{L} (\btheta)$ is essentially $l_2$-regularized square loss function and $\btheta_0$ is the randomly initialized network parameters.
However, once the sub-rewards are available, we should use different methods to train $\mathcal{F}$, in order to leverage more available information.
Next, we will elaborate our training methods using the gradient descend.

\newcommand{\newbx}{\widetilde{\mathbf{X}}_{t,k} }
\newcommand{\newrt}{\widetilde{\mathbf{r}}_{t,k} }

\para{Collection of training samples}.
Depending on the availability of sub-rewards, we apply different strategies to update $\btheta$ in each round. When the sub-rewards are all available, the learner receives one final reward and $K$ sub-rewards.  
We apply the straightforward way to train each part of $\mathcal{F}$ accordingly based on the corresponding input and the ground-truth rewards in each round, referring to the details in Algorithm \ref{alg:grad1}, where $\widetilde{m}$ in $\mathcal{F}$ should be set as $1$.

\begin{algorithm}[t]
\renewcommand{\algorithmicrequire}{\textbf{Input:}}
\renewcommand{\algorithmicensure}{\textbf{Output:}}
\caption{ $GradientDescent_{All}$ }\label{alg:grad1}
\begin{algorithmic}[1]
\Require $\mathcal{F}$, $\{ \mathbf{X}_i \}_{i=1}^t$, $ \{R_i \}_{i=1}^t $, $ \{\mathbf{r}_i \}_{i=1}^t $ ,  $J, \eta$
\Ensure $\btheta_t$
\For{each $k \in [K]$}
\State Define $\mathcal{L} (\btheta^k) =   \sum_{i=1}^t  \left(f_k(\bx_i^k; \btheta^k )-    r_i^k\right)^2 /2 + m_1\lambda\|\btheta^k - \btheta_0\|_2^2 /2 $
\For{each $j \in [J]$}
\State $\btheta^k_j = \btheta^k_{j-1} - \eta \triangledown \mathcal{L}(\btheta^k_{j-1})$
\EndFor
\EndFor
\State Define $\mathcal{L} (\bthetas) =   \sum_{i=1}^t  \left(F(\ve(\mathbf{r}_i) ; \bthetas  )-    R_i\right)^2 /2 + m_2\lambda\|\bthetas - \btheta_0\|_2^2 /2 $.  
\For{each $j \in [J]$}
\State $\bthetas_j = \bthetas_{j-1} - \eta \triangledown \mathcal{L}(\bthetas_{j-1})$
\EndFor
\State \textbf{return} $(\bthetas_J, \btheta^1_J, \dots, \btheta^K_J)$
\end{algorithmic}
\end{algorithm}

\begin{algorithm}[t]
\renewcommand{\algorithmicrequire}{\textbf{Input:}}
\renewcommand{\algorithmicensure}{\textbf{Output:}}
\caption{ $GradientDescent_{Partial}$ }\label{alg:grad2}
\begin{algorithmic}[1]
\Require $\mathcal{F}$,  $\{ \Omega_i \}_{i=1}^t$,  $J, \eta$
\Ensure $\btheta_t$
\State Define $\mathcal{L} (\btheta) =   \sum_{i=1}^t\sum_{(\mathbf{X}, R) \in \Omega_i}  \left(\mathcal{F}(\mathbf{X}; \btheta  )-    R\right)^2 /2 + m_2\lambda\|\btheta - \btheta_0\|_2^2 /2 $.  
\For{each $j \in [J]$}
\State $\btheta_j = \btheta_{j-1} - \eta \triangledown \mathcal{L}(\btheta_{j-1} )$
\EndFor
\State \textbf{return} $\btheta_j$

\end{algorithmic}
\end{algorithm}

However, when the sub-rewards are partially available, the above method is not valid anymore because the bandits without available sub-rewards cannot be trained. Therefore, to learn the $K$ bandits jointly, we propose the following training approach focusing on the empirical performance. 

As the final reward is always available in each round, we collect the first training sample $(\armc, R_t )$. 
Then, suppose there are $\mathcal{K}$ available sub-rewards $\rt$, $\mathcal{K} < K$. For each available sub-reward $r_t^k \in \rt $ and the corresponding context vector $\sarm$, we construct the following pair:
\[
\newbx = \{\mathbf{0}, \dots, \sarm, \dots, \mathbf{0}\   \}  \ \ \text{and} \ \ \newrt = \{0, \dots, r_t^k, \dots,  0\}.
\] 
We regard $\newbx$ as a new input of $\mathcal{F}$. Now, we need to determine the ground-truth final reward $\mathcal{H}(\newbx) = H(\ve(\newrt)) $.

Unfortunately, $H(\ve(\newrt))$ is unknown. Inspired by the UCB strategy, we determine $H(\ve(\newrt))$ by its upper bound. 
Based on Lemma \ref{lemma:re}, we have $H(\ve(\newrt)) \leq  \bar{C} r_t^k $. Therefore, we set $H(\ve(\newrt))$ as:
\[
H(\ve(\newrt)) = \bar{C} r_t^k
\]
because it shows the maximal potential gain for the bandit $k$.  Then, in round $t$, we can collect additional $\mathcal{K}$ sample pairs:
\[
    \{(\newbx,  \bar{C} r_t^k ) \}_{ k \in [\mathcal{K}]}.
\]
where $[\mathcal{K}]$ denotes the bandits with available sub-rewards.

Accordingly, in each round $t$, we can collect up to $\mathcal{K} + 1$ samples for training $\mathcal{F}$, denoted by $\Omega_t$,:
\begin{equation}\label{eq:armpool}
    \Omega_t  =  \{(\newbx,  \bar{C} r_t^k ) \}_{k \in \mathcal{[K]}} \bigcup \{(\armc, R_t )  \}.
\end{equation} 
Therefore, in each round,  we train $\mathcal{F}$ integrally,  based on $ \{ \Omega_i \}_{i=1}^t $, as described in Algorithm \ref{alg:grad2}.

\begin{lemma}\label{lemma:re}
Let  $\mathbf{0} = ( 0, \dots, 0 )$ and $|\mathbf{0}| = K$. Given $\newbx$ and  $\newrt$, then we have $H( \ve(\newrt)) \leq \bar{C} r_t^k$. 
\end{lemma}

\begin{prove}
As $H$ is $\bar{C}$-Lipschitz continuous, we have 
\[
| H( \ve(\newrt) ) -   H(\mathbf{0})| \leq \bar{C} \sqrt{ \sum_{r \in \newrt} (r - 0)^2} =  \bar{C}r_t^k.
\]
\end{prove}

\subsection{Upper confidence bound}

In this subsection, we present the arm selection criterion based on the upper confidence bound provided in Section 5 and then summarize the high-level idea of \sysn. 

In each round $t$, given an arm combination $\armc$, the confidence bound of $\mathcal{F}$ with respect to $\mathcal{H}$ is defined as: 
\[
\mathbb{P}\left( |\mathcal{F}(\armc; \btheta_{t}) -   \mathcal{H}(\armc)     | > \text{UCB}(\armc)   \right) \leq \delta,
\]
where UCB($\armc$) is defined in Theorem \ref{theo:ucb} and $\delta$ usually is a small constant. Then, in each round, given the all possible arm combinations $\mathbf{S}_t$, the selected $K$ arms $\armc$ are determined by: 
\begin{equation} \label{eq:cri}
\armc = \arg \max_{\mathbf{X}_t' \in \mathbf{S}_t } \left(  \mathcal{F}( 
\mathbf{X}' ; \btheta_{t}) + \text{UCB}(\armc')  \right).
\end{equation}

With this selection criterion, the workflow of \sysn is depicted in Algorithm \ref{alg:main}.

\section{Regret Analysis}
In this section, we provide the upper confidence bound and regret analysis of \sysn when the sub-rewards are all available.

Before presenting Theorem \ref{theo:ucb}, let us first focus on an $L$-layer fully-connected neural network $f(\bx_t; \btheta)$ to learn a ground-truth function $h(\bx_t)$, where $\bx \in \mathbb{R}^d$. The parameters of $f$ are set as $\mathbf{W}_1 \in \mathbb{R}^{m \times d}$,  $ \mathbf{W}_i \in \mathbb{R}^{m \times m}, \forall i \in [1 : L-1]$, and $\mathbf{W}_{L} \in \mathbb{R}^{1 \times m}$. Given the context vectors by $\trains$ and corresponding rewards $\{r_t\}_{t=1}^T$, conduct the gradient descent with the loss $\mathcal{L}$  to train $f$. 

Built upon the Neural Tangent Kernel (NTK) \cite{ntk2018neural, ntk2019generalization}, $h(\bx_t)$ can be represented by a linear function with respect to the gradient $g(\bx_t; \btheta_0)$  introduced in \cite{zhou2020neural} as the following lemma.


\newcommand{\bbxa}{\mathbf{X}_t^{\ast}}
\newcommand{\bbxt}{\mathbf{X}_t}

\begin{lemma} [Lemma 5.1 in \cite{zhou2020neural}]  \label{lemma:thetas}
There exist a positive constant $C$ such that with  probability at least $1-\delta$, if $m \geq C T^{4} L^6 \log(T^2 L/\delta)/\lambda^{4}$ for any $\bx_t \in \{\bx_t \}_{t=1}^T$, there exists a $\bts$ such that
\[
\begin{aligned}
h(\bx_t) = \langle g(\bx_t; \btheta_0),  \bts \rangle
\end{aligned}
\]
\end{lemma}  

Then, with the above linear representation of $h(\bx_t)$, we provide the following upper confidence bound with regard to $f$.

\begin{lemma} \label{lemma:iucb}
Given a set of context vectors $\{\bx_t \}_{t=1}^{T}$ and the corresponding rewards $\{r_t\}_{t=1}^{T} $ , $ \mathbb{E}(r_t) = h(\bxt)$ for any $\bxt \in \trains$. 
Let $f( \bx_t ; \btheta)$ be the $L$-layers fully-connected neural network where the width is $m$, the learning rate is $\eta$, and $\btheta \in \mathbb{R}^{P} $.  Assuming $\|\bts\|_2 \leq S/\sqrt{m} $, then, there exist positive constants $C_1, C_2$ such that if 
\[
\begin{aligned}
m &\geq \max\{ \mathcal{O} \left( T^7 \lambda^{-7} L^{21} (\log m)^3 \right ),   \mathcal{O} \left(  \lambda^{-1/2} L^{-3/2}(\log (T L^2/\delta))^{3/2}   \right) \}\\
&\eta =  \mathcal{O}( TmL +m \lambda )^{-1}, \ \   J \geq \widetilde{\mathcal{O}}(TL/\lambda),
\end{aligned} 
\]
then, with probability at least $1-\delta$, for any $\bxt \in  \{\bx_t \}_{t=1}^{T}$, we have the following upper confidence bound:
\[
\begin{aligned}
\left| \hx  - \fx     \right| \leq & \gamma_1  \| g(\bx_t; \btheta_t)/\sqrt{m} \|_{\mathbf{A}_t^{-1}}  + \gamma_2   \| g(\bx_t; \btheta_0)/\sqrt{m} \|_{\mathbf{A}_t^{' -1}} \\ 
&+  \gamma_1 \gamma_3  + \gamma_4,  \    \   \text{where}
\end{aligned}
\]

\[ 
\begin{aligned}
& \gamma_1 (m, L) =  (\lambda + t \mathcal{O}(L)) \cdot ( (1 - \eta m \lambda)^{J/2} \sqrt{t/\lambda}) +1  \\
&\gamma_2(m,L, \delta) =   \sqrt{  \log \left(  \frac{\deter(\mathbf{A}_t')} { \deter(\lambda\mathbf{I}) }   \right)   - 2 \log  \delta } + \lambda^{1/2} S \\
& \gamma_3 (m, L) = C_2 m^{-1/6} \sqrt{\log m}t^{1/6}\lambda^{-7/6}L^{7/2}\\
&\gamma_4(m, L)= C_1 m^{-1/6}  \sqrt{ \log m }t^{2/3} \lambda^{-2/3} L^3 \\
& \mathbf{A}_t =  \lambda \mathbf{I} + \sum_{i=1}^{t} g(\bx_t; \btheta_t) g(\bx_t; \btheta_t)^\intercal /m  \\
& \mathbf{A}_t' =  \lambda \mathbf{I} + \sum_{i=1}^{t} g(\bx_t; \btheta_0) g(\bx_t; \btheta_0)^\intercal /m .
\end{aligned}
\]
\end{lemma}

Now we are ready to provide an extended upper confidence bound for the proposed neural network model $\mathcal{F}$.

\begin{theorem} \label{theo:ucb}
Given the selected contexts $\{\armc\}_{t=1}^T$, the final rewards $\{R_t\}_{t=1}^T$, and all sub-rewards $\{\rt \}_{t=1}^T$, let $\mathcal{F}$ be the neural network model in \sysn.
In each round $t$, with the conditions in Lemma \ref{lemma:iucb} and suppose $\widetilde{m} =1$, then, with probability at least $1 -\delta$, for any $t \in [T] $, we have the following upper confidence bound:
\[
|\mathcal{F}(  \bbxt; \btheta_t  ) - \mathcal{H}( \bbxt )    |
\leq \bar{C} \sum_{k=1}^K \mathcal{B}^k   +  \mathcal{B}^{F} = \text{UCB}(\mathbf{X}_t),  \ \text{where}
\]
\[
\begin{aligned}
\mathcal{B}^k & = \gamma_1 \|g_k(\bx_t^k; \btheta_t^k)/\sqrt{m_1} \|_{{\mathbf{A}^k_t}^{-1}}  +  \gamma_2 (\frac{\delta}{k+1}) \|g_k(\bx_t^k; \btheta_0^k)/\sqrt{m_1} \|_{{\mathbf{A}^k_t}^{' -1}}
\\ 
& \  \  + \gamma_1 \gamma_3  + \gamma_4  \\
\mathcal{B}^F  &= \gamma_1 \| G(\mathbf{f}_t;\btheta_t^{\Sigma}) /\sqrt{m_2} \|_{{\mathbf{A}^F_t}^{-1}} + \gamma_2 (\frac{\delta}{k+1}) \| G(\mathbf{f}_t; \btheta_0^{\Sigma}) /\sqrt{m_2} \|_{{\mathbf{A}^F_t}^{'-1}} \\
& \  \  + \gamma_1 \gamma_3      + \gamma_4 \\
&\mathbf{A}^k_t =  \lambda \mathbf{I} + \sum_{i=1}^{t} g_k(\bx_i^k; \btheta_t^k) g_k(\bx_i^k; \btheta_t^k)^\intercal / m_1 \\
&\mathbf{A}^{k'}_t =  \lambda \mathbf{I} + \sum_{i=1}^{t} g_k(\bx_i^k; \btheta_0^k) g_k(\bx_i^k; \btheta_0^k)^\intercal / m_1 \\
& \mathbf{A}^F_t =   \lambda \mathbf{I} + \sum_{i=1}^{t} G(\mathbf{f}_i;\btheta_t^{\Sigma}) G(\mathbf{f}_i;\btheta_t^{\Sigma})^\intercal /m_2 \\
& \mathbf{A}^{F'}_t =   \lambda \mathbf{I} + \sum_{i=1}^{t} G(\mathbf{f}_i;\btheta_0^{\Sigma}) G(\mathbf{f}_i;\btheta_0^{\Sigma})^\intercal /m_2
\end{aligned}
\]
\end{theorem}

With the above UCB, we provide the following regret bound of \sysn.

\begin{theorem} \label{theo:reg}
Given the number of rounds $T$ and suppose that the final reward and all the sub-wards are available, let $\mathcal{F}$ be the neural network model of \sysn, satisfying the conditions in Theorem \ref{theo:ucb}. Then, assuming $\bar{C}=1$, $m_1 = m_2 =m$, $L_1 = L_2 = L$ and thus $P_1 = P_2 =P$, with probability at least $1-\delta$, the regret of \sysn is upper bounded by:
\[
\begin{aligned}
\textbf{Reg} \leq & ( \bar{C}K+1)  \sqrt{  T } 2 \sqrt{ \widetilde{P} \log ( 1 + T/\lambda)+  1/\lambda +1} \\
& \cdot \bigg(  \sqrt{ (\widetilde{P}-2) \log \left( \frac{ (\lambda + T)(1+K)}{\lambda \delta} \right) +  1/\lambda}  + \lambda^{1/2} S  +2 \bigg) + 2 (\bar{C} K+1) , \\
\end{aligned}
\]
where $\widetilde{P}$ is the effective dimension defined in Appendix (Definition \ref{def:dim}). 
\end{theorem}

\newcommand{\UCB}{\text{UCB}}

\textbf{Prove 5.4.}
First, the regret of one round $t$ :
\[
\begin{aligned}
\text{Reg}_t& = \mathcal{H}(\bbxa) - \mathcal{H}(\bbxt) \\ 
  &\leq | \mathcal{H}(\bbxa) - \mathcal{F}(\bbxa)  |  + \mathcal{F}(\bbxa) - \mathcal{H}(\bbxt) \\
  &\leq \UCB(\bbxa) + \mathcal{F}(\bbxa) - \mathcal{H}(\bbxt)\\
  &\leq  \UCB(\bbxt) + \mathcal{F}(\bbxt) - \mathcal{H}(\bbxt) \leq 2 \UCB(\bbxt) 
\end{aligned}
\]
where the third inequality is due to the selection criterion of \sysn, satisfying 
$ \mathcal{F}(\bbxa) + \UCB(\bbxa) \leq   \mathcal{F}(\bbxt) + \UCB(\bbxt).$
Thus, it has
\[
\begin{aligned}
 \textbf{Reg} &= \sum_{t =1}^T \text{Reg}_t \leq 2 \sum_{t =1}^T \UCB(\bbxt) 
\leq  2 \sum_{t =1}^T \left(  \bar{C}  \sum_{k=1}^K \mathcal{B}^k  +  \mathcal{B}^F    \right) 
\end{aligned}
\]
First, for any $k \in [K]$,  we bound 
\[
\begin{aligned}
\sum_{t=1}^{T} \mathcal{B}^k &\leq \underbrace { \gamma_1  \sum_{t=1}^{T}   \|g(\bx_t; \btheta_t) /\sqrt{m} \|^2_{ { \mathbf{A}_{t}}^{-1} } } _{\mathbf{I}_1} +   \underbrace{\gamma_2 \sum_{t=1}^{T}   \|g(\bx_t; \btheta_0) /\sqrt{m} \|^2_{ { \mathbf{A}_{t}}^{'-1} }}_{\mathbf{I}_2} \\ 
& +   T \gamma_1 \gamma_3    + T \gamma_4   \\
\end{aligned}
\]
Because the Lemma 11 in \cite{2011improved}, we have
\[
\begin{aligned}
& \mathbf{I}_1 \leq   \gamma_1 \sqrt{T \left( \sum_{t=1}^{T}   \|g(\bx_t; \btheta_t) /\sqrt{m} \|^2_{ { \mathbf{A}_{t}}^{-1} } \right)  } 
\leq  \gamma_1 \sqrt{T \left( \log\frac{\deter(\mathbf{A}_T)}{\deter(\lambda\mathbf{I}) } \right)  } \\
&\leq   \gamma_1 \sqrt{T \left(  \log\frac{\deter(\mathbf{A}_T')}{\deter\lambda\mathbf{I}) } + | \log\frac{\deter(\mathbf{A}_T)}{\deter(\lambda\mathbf{I}) } - \log\frac{\deter(\mathbf{A}_T')}{\deter(\lambda\mathbf{I}) }   | \right)  }  \\
& \leq    \gamma_1 \sqrt{T \left( \widetilde{P} \log ( 1 + T/\lambda)+  1/\lambda + 1 \right) } \\
\end{aligned}
\]
where the last inequality is based on Lemma \ref{lemma:det} and the choice of $m$. Then, applying Lemma 11 in \cite{2011improved} and  Lemma \ref{lemma:det} again, we have
\[
\begin{aligned} 
\mathbf{I}_2 & \leq  \gamma_2 \sqrt{T \left( \log\frac{\deter(\mathbf{A}_T')}{\deter(\lambda\mathbf{I}) } \right)  }\\
& \leq  \left(  \sqrt{ (\widetilde{P}-2) \log \left( \frac{ (\lambda + T)(1+K)}{\lambda \delta} \right) +  1/\lambda}  + \lambda^{1/2} S \right)\\
& \cdot \sqrt{T \left( \widetilde{P} \log ( 1 + T/\lambda)+  1/\lambda\right) }
\end{aligned}
\]
As the choice of $J$, $\gamma_1 \leq 2$. 
Then, as $m$ is sufficiently large, we have 
\[
\begin{aligned}
T  \gamma_1 \gamma_3 \leq 1,   T \gamma_4  &\leq 1.
\end{aligned}
\]
Then, because $m_1 = m_2$, $L_1 = L_2$ and $\bar{C}=1$,      we have
\[
\textbf{Reg} \leq 2(\bar{C} K+1) \sum_{t=1}^{T} \mathcal{B}^k.
\]
Putting everything together proves the claim.
\newline 

$\widetilde{P}$ is the effective dimension defined by the eigenvalues of the NTK ( Definition \ref{def:dim} in Appendix). Effective dimension was first introduced by \cite{valko2013finite} to analyze the kernelized context bandit, and then was extended to analyze the kernel-based Q-learning\cite{yang2020reinforcement} and the neural-network-based bandit \cite{zhou2020neural}. 
$\widetilde{P}$ can be much smaller than the real dimension $P$, which alleviates the predicament when $P$ is extremely large.

Theorem \ref{theo:reg} provides the $\widetilde{\mathcal{O}}\left((K+1) \sqrt{T} \right)$ regret bound for \sysn, achieving the near-optimal bound compared with  a single bandit (  $\widetilde{\mathcal{O}}(\sqrt{T})$   )  that is either linear \cite{2011improved} or non-linear \cite{valko2013finite,zhou2020neural}. With different width $m_1, m_2$ and the Lipschitz continuity $\bar{C}$, the regret bound of \sysn becomes $\widetilde{\mathcal{O}}( ( \bar{C}K + 1) \sqrt{T})$.

\section{experiments}

To evaluate the empirical performance of  \sysn, in this section, we design two different multi-facet bandit problems on three real-world data sets.
The experiments are divided into two parts to evaluate the effects of final rewards and availability of sub-rewards. The code has been released \footnote{https://github.com/banyikun/KDD2021\_MuFasa}.

\para{Recommendation:Yelp\footnote{https://www.yelp.com/dataset}}.
Yelp is a data set released in the Yelp data set challenge, which consists of 4.7 million rating entries for  $1.57 \times 10^5$ restaurants by $1.18$ million users. In addition to the features of restaurants, this data set also provides the attributes of each user and the list of his/her friends. 
In this data set, we evaluate \sysn on personalized recommendation, where the learner needs to simultaneously recommend a restaurant and a friend (user) to a served user. Naturally, this problem can be formulated into $2$ bandits in which one set of arms $\mathbf{X}_t^1$ represent the candidate restaurants and the other set of arms $\mathbf{X}_t^2$ formulates the pool of friends for the recommendation.
We apply LocallyLinearEmbedding\cite{roweis2000nonlinear} to train a $10$-dimensional feature vector $\bx_{t,i}^{1}$ for each restaurant and a $6$-dimensional feature vector $\bx_{t,j}^{2}$ for each user.  
Then, for the restaurant, we define the reward according to the rating star: The reward $r_t^1$ is $1$ if the number of rating stars is more than 3 (5 in total); Otherwise, the reward $r_t^1$ is $0$. For friends, the reward $r_t^2 = 1$ if the recommended friend is included in the friend list of the served user in fact; Otherwise  $r_t^2 = 0$. 
To build the arm sets, we extract the rating entries and friends lists of top-10 users with the most ratings. In each round $t$, we build the arm set $\mathbf{X}_t^1$ and $\mathbf{X}_t^2$ by picking one restaurant/friend with $1$ reward and then randomly picking the other $9$ restaurants/friends with $0$ rewards. 
Thus $|\mathbf{X}_t^1| = |\mathbf{X}_t^2| = 10$.

\para{Classification:Mnist \cite{lecun1998gradient} + NotMnist}.
These are two well-known 10-class classification data sets.
The evaluation of contextual bandit has been adapted to the classification problem \cite{zhou2020neural, deshmukh2017multi,valko2013finite}. Therefore, we utilize these two similar classification data sets to construct $2$ bandits, where the 10-class classification is converted into a $10$-armed contextual bandit.
Considering a sample figure $\bx \in \mathbb{R}^{d}$, we tend to classify it from $10$ classes. Under the contextual bandit setting, $\bx$ is transformed into $10$ arms: $\bx_{1} = (\bx, \mathbf{0}, \dots, \mathbf{0});  \bx_{2} = ( \mathbf{0}, \bx , \dots, \mathbf{0}); \dots; \bx_{10} = ( \mathbf{0},  \mathbf{0} , \dots, \bx) \in \mathbb{R}^{10d}$ matching the $10$ classes. In consequence, the reward is $1$ if the learner plays the arm that matches the real class of $\bx$; Otherwise, the reward is $0$. Using this way, we can construct two contextual bandits for these two data sets, denoted by $(\mathbf{X}_t^1, r_t^1)$ and  $(\mathbf{X}_t^2, r_t^2)$.
Then, in each round, the arm pools will be $|\mathbf{X}_t^1| = |\mathbf{X}_t^2| = 10$.

To evaluate the effect of different final reward function, with the sub-rewards $\rt = \{r_t^1, r_t^2\}$, we design the following final reward function:
\begin{equation} \label{eq:h12}
 H_1(\ve(\rt)) =  r_t^1 + r_t^2;  \  \ H_2(\ve(\rt)) =  2r_t^1 + r_t^2.  \\
\end{equation}
For (1), it describes the task where each bandit contributes equally. Of (2), it represents some tasks where each bandit has different importance.

As the problem setting is new, there are no existing algorithms that can directly adapt to this problem. Therefore, we construct baselines by extending the bandit algorithms that work on  a single bandit, as follows:

\begin{enumerate}
    \item \textbf{(K-)LinUCB}. LinUCB \cite{2010contextual} is a linear contextual bandit algorithm where the reward function is assumed as the dot product of the arm feature vector and an unknown user parameter. Then, apply the UCB strategy to select an arm in each round. To adapt to the multi-facet bandit problem, we duplicate LinUCB for $K$ bandits. For example, in Yelp data set, we use two LinUCB to recommend restaurants and friends, respectively. 
    \item \textbf{(K-)KerUCB }. KerUCB \cite{valko2013finite} makes use of a predefined kernel matrix to learn the reward function and then build a UCB for exploration. We replicate $K$ KerUCB to adapt to this problem.
    \item \textbf{(K-)NeuUCB}. NeuUCB\cite{zhou2020neural} uses a fully-connected neural network to learn one reward function with the UCB strategy. Similarly, we duplicate it to $K$ bandits.
\end{enumerate}

\para{Configurations}. 
 For \sysn, each sub-network $f_k(\bx_t^k; \btheta^k)$ is set as a two-layer network:  $f_k(\bx_t^k; \btheta^k) = \sqrt{m_1} \mathbf{W}_2 \sigma(\mathbf{W}_1 \bxtk)$, where $\mathbf{W}_1 \in \mathbb{R}^{m_1\times d_k},   \mathbf{W}_2 \in \mathbb{R}^{\widetilde{m} \times m_1}$, and $m_1 = \widetilde{m}= 100$.
 Then, the shared layers $F(\mathbf{f}_t;\bthetas) = \sqrt{m_2} \mathbf{W}_2 \sigma(\mathbf{W}_1 \mathbf{f}_t)$, where $\mathbf{W}_1 \in \mathbb{R}^{m_2 \times 2\widetilde{m} },   \mathbf{W}_2 \in \mathbb{R}^{1 \times m_2}$ and $m_2 = 100$. 
For the $H_1$, $\bar{C}$ is set as $1$ and set as $2$ for $H_2$.
To learn $K$ bandits jointly, in the experiments, we evaluate the performance of Algorithm $1 + 3$.
For K-NeuUCB, for each NeuUCB, we set it as a $4$-layer fully-connected network with the same width $m=100$ for the fair comparison. The learning rate $\eta$ is set as $0.01$ and the upper bound of ground-truth parameter $S = 1$ for these two methods.
To accelerate the training process, we update the parameters of the neural networks every $50$ rounds.
For the KerUCB, we use the radial basis function (RBF) kernel and stop adding contexts to KerUCB after 1000 rounds, following the same setting for Gaussian Process in \cite{riquelme2018deep, zhou2020neural}. 
For all the methods, the confidence level $\delta = 0.1$, the regularization parameter $\lambda = 1$. All experiments are repeatedly run $5$ times and report the average results.

\subsection{Result 1: All sub-rewards with different $H$}

\begin{figure}[t] 
    \includegraphics[width=1.0\columnwidth]{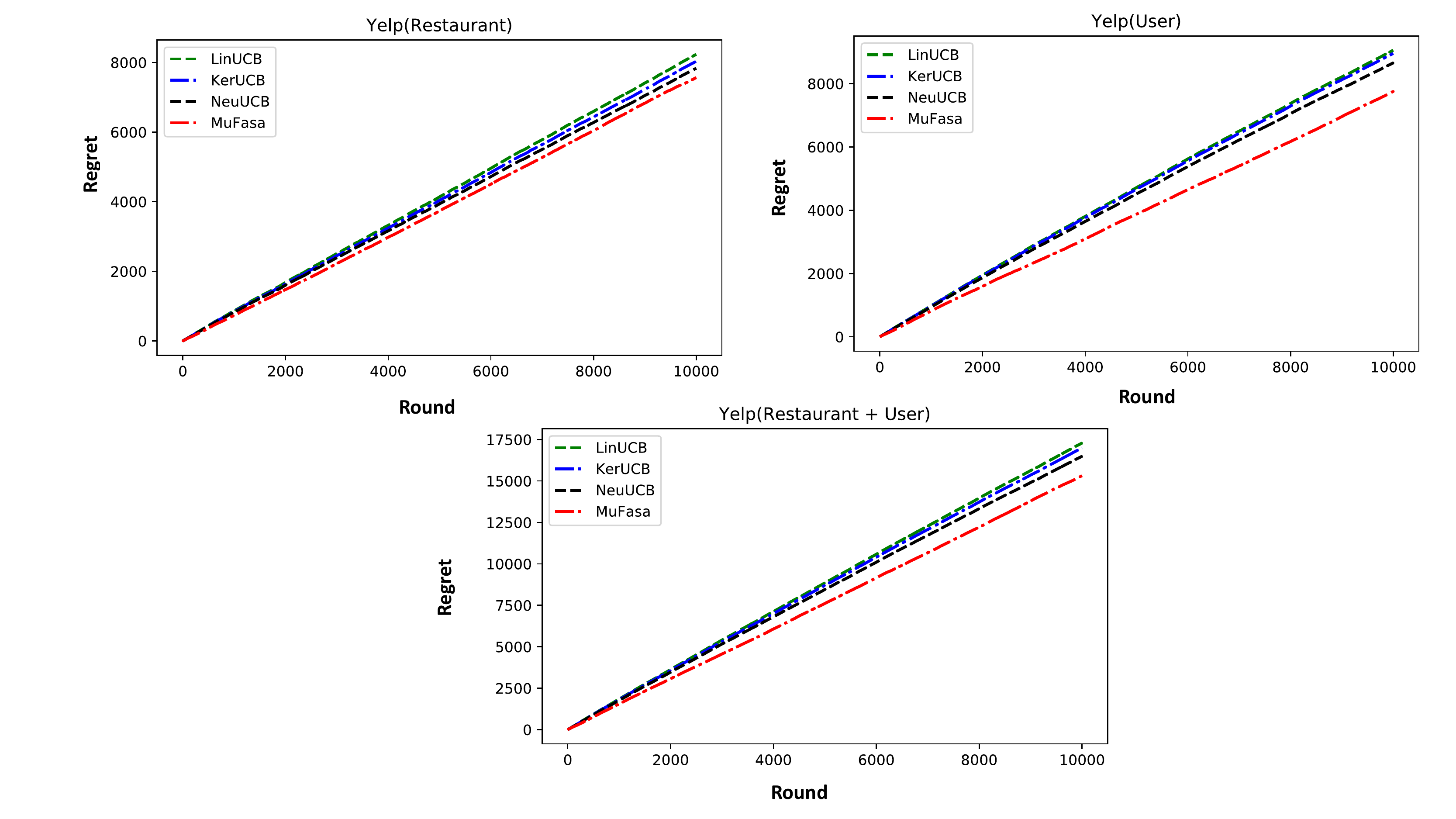}
    \centering
    \caption{Regret comparison on Yelp with $H_1$ final reward function.}
       \label{fig:yelp_f1}
\end{figure}

\begin{figure}[t] 
    \includegraphics[width=1.0\columnwidth]{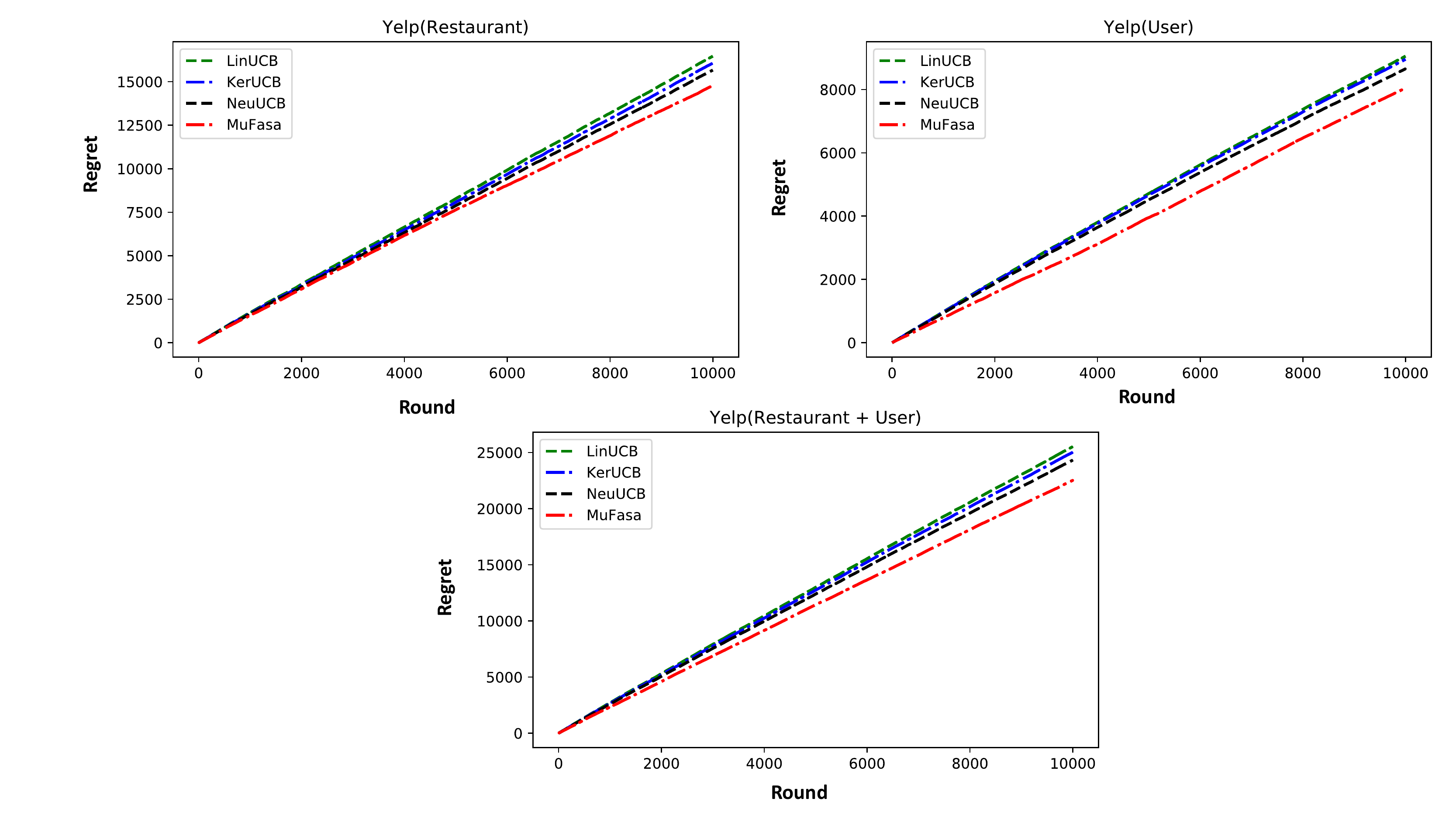}
    \centering
    \caption{Regret comparison on Yelp with $H_2$ final reward function.}
       \label{fig:yelp_f2}
\end{figure}

\begin{figure}[t] 
    \includegraphics[width=0.9\columnwidth]{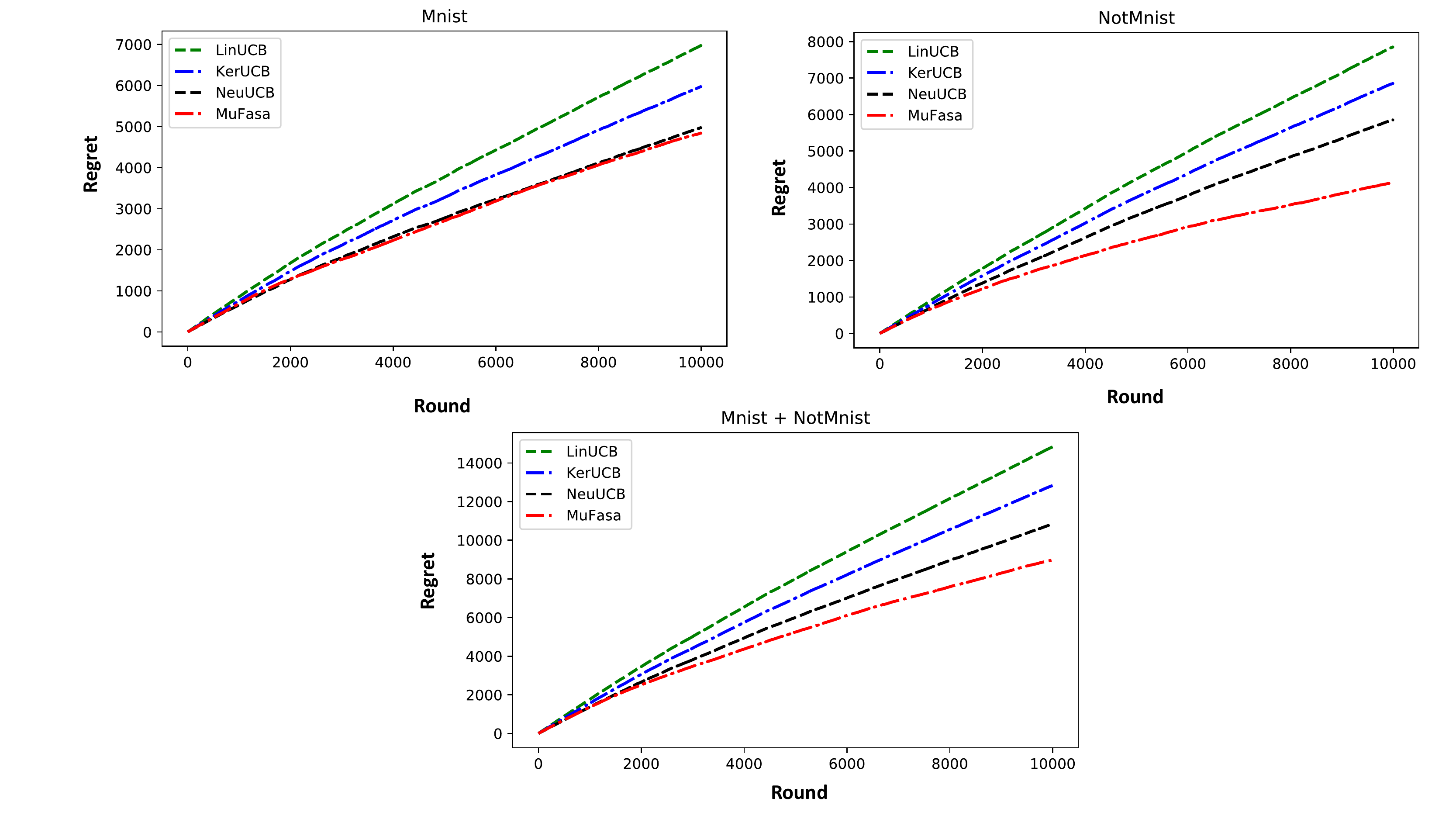}
    \centering
    \vspace{-1em}
    \caption{Regret comparison on Mnist+NotMnist with $H_1$.}
       \label{fig:mnist_f1}
\end{figure}

With the final reward function  $H_1$  (Eq.(\ref{eq:h12})), 
Figure \ref{fig:yelp_f1} and Figure \ref{fig:mnist_f1} report the regret of all methods on Yelp and Mnist+NotMnist data sets, where the first top-two sub-figure shows the regret of each bandit and the bottom sub-figure shows the regret of the whole task (2 bandits). 
These figures show that \sysn achieves the best performance (the smallest regret), because it utilizes a comprehensive upper confidence bound built on the assembled neural networks to select two arms jointly in each round. 
This indicates the good performance of \sysn on personalized recommendation and classification. 
Among these baselines, NeuUCB achieves the best performance, which thanks to the representation power of neural networks. 
However, it chooses each arm separately, neglecting the collaborative relation of $K$ bandits. For KerUCB, it shows the limitation of the simple kernels like the radial basis function compared to neural network. LinUCB fails to handle each task, as it assume a linear reward function and thus usually cannot to learn the complicated reward functions in practice.

\begin{figure}[t] 
    \includegraphics[width=0.9\columnwidth]{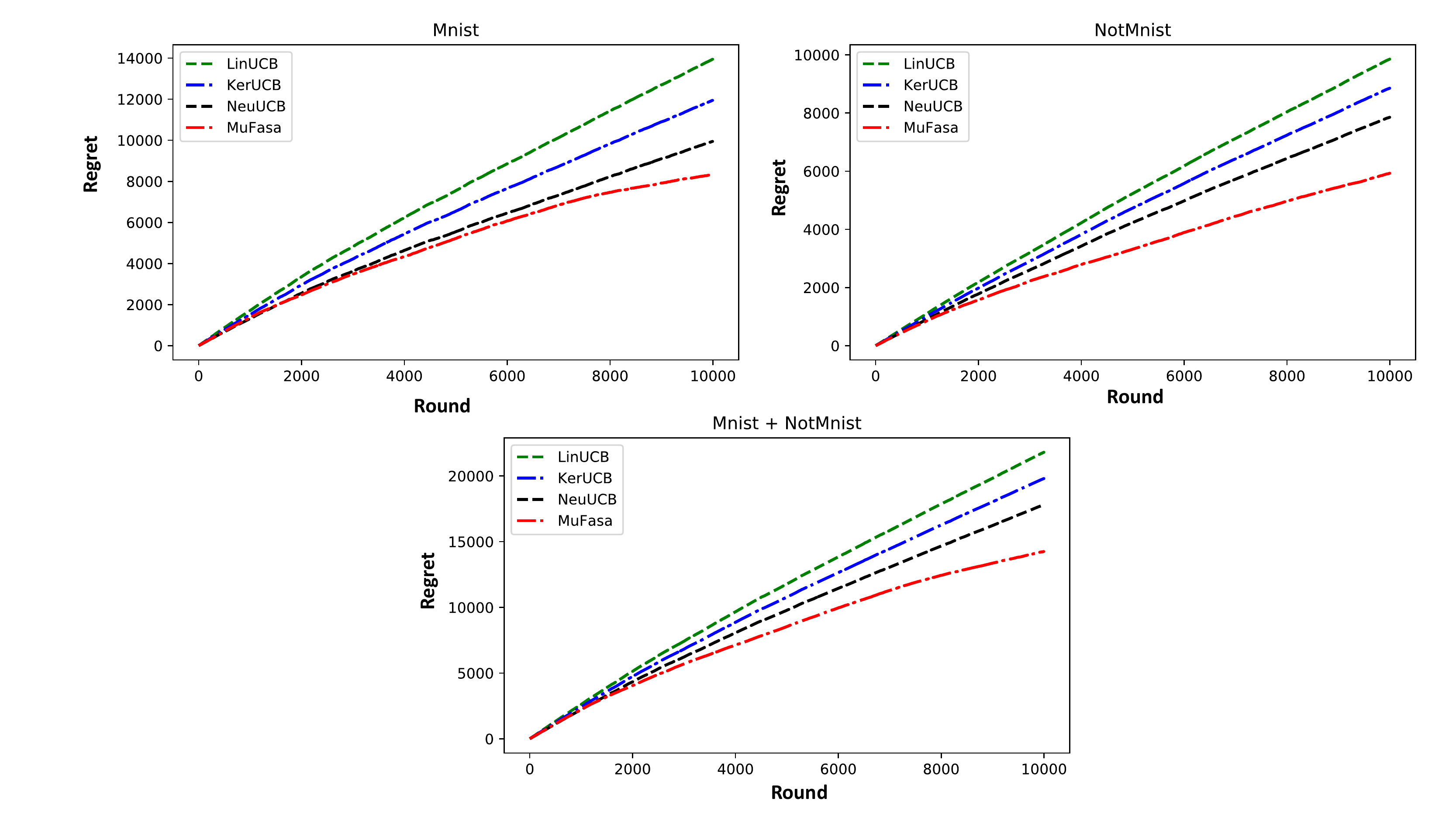}
    \centering
    \vspace{-1em}
    \caption{Regret comparison on Mnist+NotMnist with $H_2$.}
       \label{fig:mnist_f2}
\end{figure}

With the final reward function  $H_2$ (Eq.(\ref{eq:h12})), 
Figure \ref{fig:yelp_f2} and Figure \ref{fig:mnist_f2} depict the regret comparison on Yelp and Mnist+NotMnist data sets. 
The final reward function $H_2$ indicates that the bandit 1 weights more than bandit 2 in the task. 
Therefore, to minimize the regret, the algorithm should place the bandit 1 as the priority when making decisions. As the design of \sysn, the neural network $\mathcal{F}$ can learn the relation among the bandits.    
For example, on the Mnist and NotMnist data sets, consider two optional select arm sets $ \{x_{t, i_1}^1, x_{t,i_2}^2\}$ and  $\{x_{t, j_1}^1, x_{t, j_2}^2 \}$. 
The first selected arm set receives $1$ reward on Mnist while $0$ reward on NotMnist. In contrast, the second selected arm set receives $0$ reward on Mnist while $1$ reward on NotMnist. 
However, these two bandits have different weights and thus the two arm sets have different final rewards, i.e., $R_t^1 = 2 $ and  $R_t^2  = 1$, respectively.
To maximize the final reward, the learner should select the first arm set instead of the second arm set. 
As \sysn can learn the weights of bandits, it will give more weight to the first bandit and thus select the first arm set. 
On the contrary, all the baselines treat each bandit equally, and thus they will select these two arm sets randomly.
Therefore, under the setting of $H_2$, with this advantage, \sysn further decreases the regret on both Yelp and Mnist+NotMnist data sets. For instance, on Mnist+NotMnist data sets, \sysn with $H_2$ decrease $20 \%$ regret over NeuUCB while  \sysn with $H_1$ decrease $17.8 \%$ regret over NeuUCB.

\subsection{Result 2: Partial sub-rewards}

As the sub-rewards are not always available in many cases, in this subsection, we evaluate \sysn with partially available sub-rewards on Yelp and Mnist+NotMnist data sets. 
Therefore, we build another two variants of \sysn: (1) \sysn(One sub-reward) is provided with the final reward and one sub-reward of the first bandit; (2) \sysn(No sub-reward) does not receive any sub-rewards except the final reward. Here, we use the $H_1$ as the final reward function.

\begin{figure}[t] 
    \includegraphics[width=0.9\columnwidth]{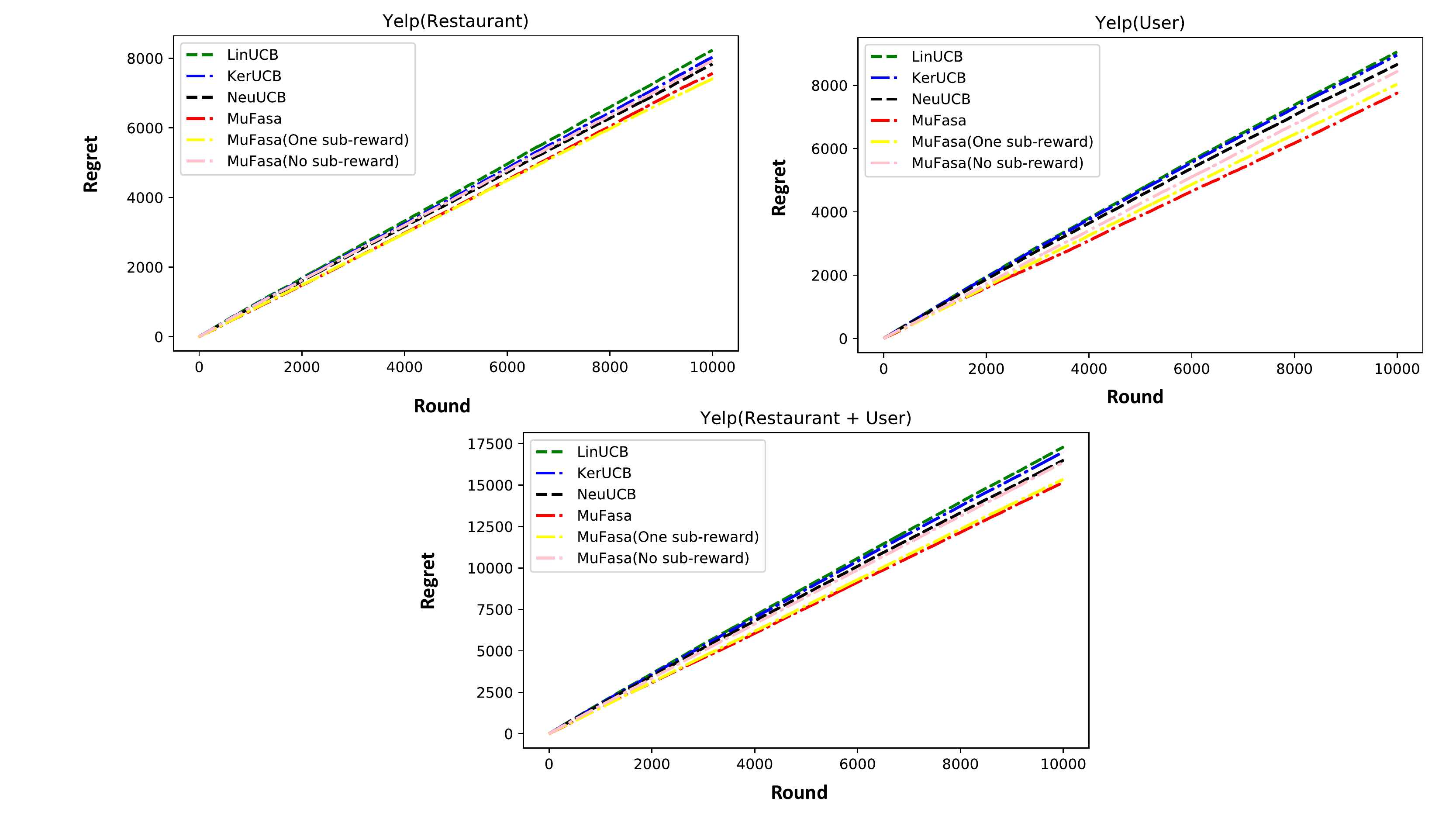}
    \centering
    \caption{Regret comparison on Yelp with different reward availability.}
       \label{fig:yelp_one}
\end{figure}

\begin{figure}[t] 
    \includegraphics[width=0.9\columnwidth]{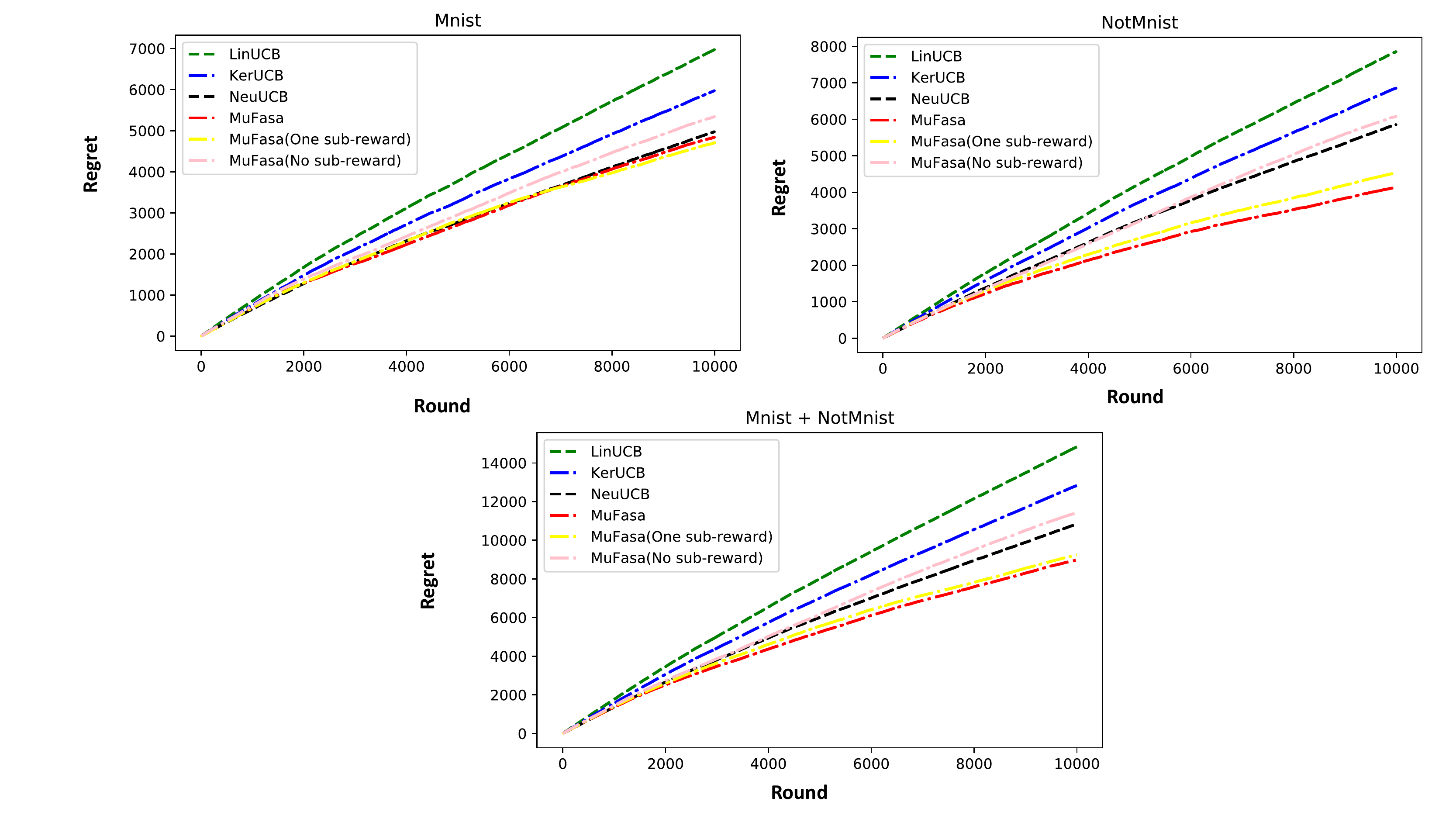}
    \centering
    \caption{Regret comparison on Mnist+NotMnist  with different reward availability.}
       \label{fig:mnist_one}
\end{figure}

Figure \ref{fig:yelp_one} and Figure \ref{fig:mnist_one} show the regret comparison with the two variants of \sysn, where \sysn exhibits the robustness with respect to the lack of sub-rewards. 
Indeed, the sub-reward can provide more information to learn, while \sysn(One sub-reward) still outperforms all the baselines, because the final reward enables \sysn to learn the all bandits jointly and the sub-reward strengthens the capacity of learning the exclusive features of each bandit. 
In contrast, all the baselines treat each bandit separately.
Without any-rewards, \sysn still achieves the acceptable performance. On the Yelp data set, the regret of \sysn(No sub-reward) is still lower than the best baseline NeuUCB while lacking considerable information. On the Mnist+NotMnist data set, although \sysn(No sub-reward) does not outperform the baselines, its performance is still closed to NeuUCB. Therefore, as long as the final reward is provided, \sysn can tackle the multi-facet problem effectively. Moreover, \sysn can leverage available sub-rewards to improve the performance.

\section{Conclusion}

In this paper, we define and study the novel problem of the multi-facet contextual bandits, motivated by real applications such as comprehensive personalized recommendation and healthcare. We propose a new bandit algorithm, \sysn. 
It utilizes the neural networks to learn the reward functions of multiple bandits jointly and explores new information by a comprehensive upper confidence bound. Moreover, we prove that \sysn can achieve the $\widetilde{ \mathcal{O}} ((K+1)\sqrt{T})$ regret bound under mild assumptions. Finally, we conduct extensive experiments to show the effectiveness of \sysn\ on personalized recommendation and classification tasks, as well as the robustness of \sysn\ in the lack of sub-rewards.

\section*{Acknowledgement}
This work is supported by National Science Foundation under Award No. IIS-1947203 and IIS-2002540, and the U.S. Department of Homeland Security under Grant Award Number 17STQAC00001-03-03. The views and conclusions are those of the authors and should not be interpreted as representing the official policies of the funding agencies or the government.

\bibliographystyle{ACM-Reference-Format}
\bibliography{ref}

\clearpage

\section{appendix}

\begin{definition} \label{def:ridge}
Given the context vectors $\{\bx_t\}_{t=1}^T$ and the rewards $\{ r_t \}_{t=1}^{T} $, then we define the estimation $\btheta'$ via ridge regression:  
\[
\begin{aligned}
&\mathbf{A}_t' =  \lambda \mathbf{I} + \sum_{i=1}^{t} g(\bx_t; \btheta_0) g(\bx_t; \btheta_0)^\intercal /m\\
&\mathbf{b}_t' = \sum_{i=1}^t r_t g(\bx_t; \btheta_0)/\sqrt{m}  \\
&\btheta' = \mathbf{A}^{-1}_t\mathbf{b}_t  \\
& \mathbf{A}_t =  \lambda \mathbf{I} + \sum_{i=1}^{t} g(\bx_t; \btheta_t)  g(\bx_t; \btheta_t)^\intercal /m
\end{aligned}
\]
\end{definition}

\begin{definition} [ NTK \cite{ntk2018neural, ntk2019generalization}] Let $\mathcal{N}$ denote the normal distribution.
Define 
\[
\begin{aligned}
&\mathbf{M}_{i,j}^0 = \bsigma^{0}_{i,j} =  \langle \bx_i, \bx_j\rangle,   \ \ 
\mathbf{N}^{l}_{i,j} =
\begin{pmatrix}
\bsigma^{l}_{i,i} & \bsigma^{l}_{i,j} \\
\bsigma^{l}_{j,i} &  \bsigma^{l}_{j,j} 
\end{pmatrix} \\
&   \bsigma^{l}_{i,j} = 2 \mathbb{E}_{a, b \sim  \mathcal{N}(\mathbf{0}, \mathbf{N}_{i,j}^{l-1})}[ \sigma(a) \sigma(b)] \\
& \mathbf{M}_{i,j}^l = 2 \mathbf{M}_{i,j}^{l-1} \mathbb{E}_{a, b \sim \mathcal{N}(\mathbf{0}, \mathbf{N}_{i,j}^{l-1})}[ \sigma'(a) \sigma'(b)]  + \bsigma^{l}_{i,j}.
\end{aligned}
\]
Then, given the contexts $\{\bx_t\}_{t=1}^T$, the Neural Tangent Kernel is defined as $ \mathbf{M} =  (\mathbf{M}^L + \bsigma^{L})/2$.
\end{definition}

\begin{definition} [Effective Dimension  \cite{zhou2020neural}] \label{def:dim}
Given the contexts $\{\bx_t\}_{t=1}^T$, the effective dimension $\widetilde{P}$ is defined as
\[
 \widetilde{P} = \frac{\log \deter(\mathbf{I} + \mathbf{M}/\lambda)}{ \log(1 + T/\lambda)}. 
\]
\end{definition}

\textbf{Proof of Lemma \ref{lemma:iucb}}.
Given a set of context vectors $\{\bx\}_{t=1}^{T}$ with the ground-truth function $h$ and a fully-connected neural network $f$, we have
\[
\begin{aligned}
 &\left| \hx  - \fx      \right| \\
\leq &  \left  | \hx  - \langle \gx, \btheta' /\sqrt{m} \rangle \right|  + \left| \fx -  \langle \gx, \btheta'/\sqrt{m} \rangle  \right|
\end{aligned}
\]
where $\btheta'$ is the estimation of ridge regression from Definition \ref{def:ridge}. Then, based on the Lemma \ref{lemma:thetas}, there exists $\bts \in \mathbf{R}^{P}$ such that $ h(\bxt) =  \left \langle  g(\bx_i, \btheta_0), \bts \right \rangle$. Thus, we have
\[
\begin{aligned}
& \ \   \left  | \hx  - \langle \gx, \btheta'/\sqrt{m} \rangle \right| \\
 &  =  \left|  \left \langle  g(\bx_i, \btheta_0)/\sqrt{m},  \sqrt{m} \bts \right \rangle   -   \left \langle  g(\bx_i, \btheta_0)/\sqrt{m},  \btheta' \right \rangle \right |
\leq  \\
& \left ( \sqrt{  \log \left(  \frac{\deter(\mathbf{A}_t')} { \deter(\lambda\mathbf{I}) }   \right)   - 2 \log  \delta }  + \lambda^{1/2} S   \right) \| \gx/\sqrt{m}  \|_{\mathbf{A}_t^{' -1}}
\end{aligned}
\]
where the final inequality is based on the the Theorem 2 in \cite{2011improved}, with probability at least $1-\delta$, for any $t \in [T]$.

Second we need to bound 
\[
\begin{aligned}
&\left| \fx -  \langle g(\bx_t; \btheta_0), \btheta'/\sqrt{m} \rangle  \right| \\
 \leq & \left |  \fx - \langle g(\bx_t; \btheta_0), \btheta_t - \btheta_0 \rangle   \right|  \\
 &+    \left|     \langle g(\bx_t; \btheta_0), \btheta_t - \btheta_0 \rangle   - \langle g(\bx_t; \btheta_0), \btheta'/\sqrt{m} \rangle \right|    
\end{aligned} 
\] 

To bound the above inequality,  we first bound
\[
\begin{aligned}
 & \left |  \fx - \langle g(\bx_t; \btheta_0), \btheta_t - \btheta_0 \rangle   \right| \\
=& \left |  \fx - f(\mathbf{x}_t; \btheta_0)   - \langle g(\bx_t; \btheta_0), \btheta_t - \btheta_0 \rangle   \right| \\
\leq  & C_2 \tau^{4/3} L^3 \sqrt{m \log m} 
\leq   C_2 m^{-1/6}  \sqrt{ \log m }t^{2/3} \lambda^{-2/3} L^3.  \\
\end{aligned}
\] 
where $ f(\mathbf{x}_t; \btheta_0) = 0$ due to the random initialization of $\btheta_0$. The first inequality is derived by Lemma \ref{lemma:f}. According to the Lemma \ref{lemma:b2}, it has $\|\btheta_t - \btheta_0 \|_2 \leq 2\sqrt{\frac{t}{m\lambda}} $. Then,  replacing $\tau$ by $2\sqrt{\frac{t}{m\lambda}}$, we obtain the second inequality.

Next, we need to bound
\[ 
\begin{aligned}
 &|  \langle g(\bx_t; \btheta_0), \btheta_t - \btheta_0 \rangle - \langle g(\bx_t; \btheta_0), \btheta'/\sqrt{m} \rangle | \\
 = & |\langle g(\bx_t; \btheta_0) /\sqrt{m},     \sqrt{m} (\btheta_t - \btheta_0 -  \btheta'/\sqrt{m}) \rangle|  \\ 
\leq & \| g(\bx_t; \btheta_0) /\sqrt{m}\|_{\mathbf{A}_t^{-1}} \cdot  \sqrt{m} \| \btheta_t - \btheta_0  - \btheta'/\sqrt{m}\|_{\mathbf{A}_t} \\
\leq & \| g(\bx_t; \btheta_0) / \sqrt{m} \|_{\mathbf{A}_t^{-1}} \cdot \sqrt{m}  \|{\mathbf{A}_t} \|_2  \cdot  \| \btheta_t - \btheta_0  - \btheta'/\sqrt{m}\|_2. \\
\end{aligned} 
\]
Due to the Lemma \ref{lemma:g} and Lemma \ref{lemma:b2}, we have
\[
\begin{aligned}
 &  \sqrt{m}  \|{\mathbf{A}_t} \|_2 \cdot   \| \btheta_t - \btheta_0  - \btheta'/\sqrt{m}\|_2 \leq \sqrt{m} (\lambda + t \mathcal{O}(L)) \\
   &\cdot \left(   (1 - \eta m \lambda)^{J/2} \sqrt{t/(m\lambda)} + C_4 m^{-2/3} \sqrt{\log m}L^{7/2}t^{5/3} \lambda^{-5/3}(1+\sqrt{t/\lambda})   \right)  \\
  & \leq (\lambda + t \mathcal{O}(L)) \\
&  \cdot ( (1 - \eta m \lambda)^{J/2} \sqrt{t/\lambda} +  C_4 m^{-1/6} \sqrt{\log m}L^{7/2}t^{5/3} \lambda^{-5/3}(1+\sqrt{t/\lambda})  ) \\
   & \leq (\lambda + t \mathcal{O}(L)) \cdot ( (1 - \eta m \lambda)^{J/2} \sqrt{t/\lambda}) + 1
 \end{aligned}
\] 
where the last inequality is because $m$ is sufficiently large.
Therefore, we have 
\[
\begin{aligned}
&\left| \fx -  \langle \gx, \btheta'/\sqrt{m} \rangle  \right| \\
\leq &  \left( (\lambda + t \mathcal{O}(L)) \cdot ( (1 - \eta m \lambda)^{J/2} \sqrt{t/\lambda}) +1\right)     \| \gx / \sqrt{m} \|_{\mathbf{A}_t^{-1}}  \\
 & +  C_2 m^{-1/6}  \sqrt{ \log m }t^{2/3} \lambda^{-2/3} L^3  \\
\end{aligned} 
\] 

And we have 
\[
\begin{aligned} 
& \| \gx/\sqrt{m}  \|_{\mathbf{A}_t^{-1}} \\
 =& \| g(\bx_t; \btheta_t)  +  g(\bx_t; \btheta_0)    -   g(\bx_t; \btheta_t)      \|_{\mathbf{A}_t^{-1}} /\sqrt{m} \\
 \leq & \| g(\bx_t; \btheta_t)  /\sqrt{m} \|_{\mathbf{A}_t^{-1}} +  \| \mathbf{A}_t^{-1}\|_2  \|   g(\bx_t; \btheta_0)   -   g(\bx_t; \btheta_t)  \|_2/\sqrt{m} \\
 \leq  & \| g(\bx_t; \btheta_t)  /\sqrt{m} \|_{\mathbf{A}_t^{-1}} + \lambda^{-1} m^{-1/6} \sqrt{\log m}t^{1/6}\lambda^{-1/6}L^{7/2}
\end{aligned}
\]
where the last inequality is because of Lemma \ref{lemma:b3} with Lemma \ref{lemma:b2} and $\| \mathbf{A}_t \|_2 \geq \| \lambda \mathbf{I}\|_2$.

Finally, putting everything together, we have
\[
\begin{aligned}
\left| \hx  - \fx     \right| &\leq \gamma_1  \| g(\bx_t; \btheta_t)/\sqrt{m} \|_{\mathbf{A}_t^{-1}}  + \gamma_2   \| g(\bx_t; \btheta_0)/\sqrt{m} \|_{\mathbf{A}_t^{' -1}}  \\
& +  \gamma_1 \gamma_3  + \gamma_4.
\end{aligned}
\]
\newline

\textbf{Proof of Theorem \ref{theo:ucb}}.
First, considering an individual bandit $k \in [K]$ with the set of context vectors $\{ \bx_t^k \}_{t=1}^{T}$ and the set of sub-rewards $ \{r_t^k\}_{t=1}^{T}$, we can build upper confidence bound of $f_k(\bx_t^k; \btheta^k_t)$ with respect to $h_k(\bxtk)$ based on the Lemma \ref{lemma:iucb}. Denote the UCB by $\mathcal{B}(\bxtk, m, L_1, \delta')$, with probability at least $1-\delta'$, for any $t \in [T]$  we have
\[
|f_k(\bx_t^k; \btheta^k_t) - h_k(\bxtk)| \leq   \mathcal{B}(\bxtk, m, L_1, \delta', t) = \mathcal{B}^k.
\]
Next, apply the union bound on the $K+1$ networks, we have $ \delta' = \delta/(K+1) $ in each round $t$.

Next we need to bound
\[ 
\begin{aligned}
&\left |H \left ( \ve(\rt)  \right)  -   H\left (  \mathbf{f}_t  \right) \right| \\
 \leq &\bar{C} \sqrt{ \sum_{k=1}^K  |f_k(\bx_t^k; \btheta^k_t) - h_k(\bxtk)|^2} \\ 
\leq & \bar{C} \sqrt{ \sum_{k=1}^K (\mathcal{B}^k)^2 }  \leq  \bar{C} \sum_{k=1}^K \mathcal{B}^k
\end{aligned}
\] 
where the first inequality is because $H$ is a $\bar{C}$-lipschitz continuous function. Therefore, we have 
\[
\begin{aligned}
&|\mathcal{F}( \armc ) - \mathcal{H}( \armc  )    | = | F(\mathbf{f}_t; \bthetas) -  H (   \ve(\rt)    )   |  \\
\leq & \left |F(\mathbf{f}_t; \bthetas) -    H\left ( \mathbf{f}_t \right) \right | +  \left |   H\left (  \mathbf{f}_t  \right) -  H ( \ve(\rt)  ) \right | 
\leq    \bar{C} \sum_{k=1}^K \mathcal{B}^k  +  \mathcal{B}^{F}.
\end{aligned}
\]
This completes the proof of the claim.
\newline

\begin{lemma} \label{lemma:det}
With probability at least $1-\delta'$, 
we have
\[
\begin{aligned}
(1) \|\mathbf{A}_t\|_2,  \|\mathbf{A}_t'\|_2  & \leq \lambda + t\mathcal{O}(L) \\
(2) \log \frac{\deter(\mathbf{A}_t') }{\deter( \lambda \mathbf{I})}  & \leq  \widetilde{P} \log ( 1 + T/\lambda)+  1/\lambda \\
(3) | \log \frac{\deter(\mathbf{A}_t) }{\deter( \lambda \mathbf{I})}  - \log \frac{\deter(\mathbf{A}_t') }{\deter( \lambda \mathbf{I})}| &\leq \mathcal{O}(m^{-1/6} \sqrt{\log m } L^4 t^{5/3} \lambda^{-1/6}).
\end{aligned}
\]
where $(3)$ is referred from  Lemma B.3 in \cite{zhou2020neural}.
\end{lemma}  

\textbf{Proof of Lemma \ref{lemma:det}}.
For (1), based on the Lemma \ref{lemma:g}, for any $\bxt \in \trains $, \\
$ \|\gx \|_F \leq \mathcal{O}(\sqrt{mL})$.
Then, for the first item:
\[
\begin{aligned}
&\|  \mathbf{A}_t \|_2  =  \| \lambda \mathbf{I} + \sum_{i=1}^{t} g(\bx_i; \btheta_t) g(\bx_i; \btheta_t)^\intercal /m \|_2 \\
& \leq   \| \lambda \mathbf{I} \|_2 + \| \sum_{i=1}^{t} g(\bx_i; \btheta_t) g(\bx_i; \btheta_t)^\intercal /m \|_2  \\ 
&\leq  \lambda  +  \sum_{i=1}^{t} \| g(\bx_i; \btheta_t) \|_2^2/m \leq  \lambda  +  \sum_{i=1}^{t} \|g(\bx_i; \btheta_t) \|_F^2/m  \\
&\leq \lambda + t\mathcal{O}(L).
\end{aligned}
\]
Same proof workflow for $\|  \mathbf{A}_t' \|_2$.
For (2), we have
\[
\begin{aligned}
\log \frac{\deter(\mathbf{A}_t') }{\deter( \lambda \mathbf{I})} &= \log \deter(\mathbf{I} + \sum_{t=1}^{T} \gx \gx^{\intercal}/ (m \lambda)  ) \\
& = \deter( \mathbf{I} + \mathbf{G} \mathbf{G}^{\intercal} /\lambda) 
\end{aligned}
\]
where $\mathbf{G} = ( g(\bx_1; \btheta_0), \dots, g(\bx_T; \btheta_0))/\sqrt{m}$.

According to the Theorem 3.1 in \cite{arora2019exact}, when $m = \Omega(\frac{L^6\log{L/\delta}}{\epsilon^4}) $, with probability at least $1-\delta$, for any $\bx_i, \bx_j \in \{\bx_t\}_{t=1}^T$, it has 
\[
|g(\bx_i; \btheta_0)^{\intercal} g(\bx_j; \btheta_0)/m - \mathbf{M}_{i,j}| \leq \epsilon.
\]
Therefore, we have 
\[
\begin{aligned}
\|  \mathbf{G} \mathbf{G}^{\intercal} - \mathbf{M}\|_F  & =\sqrt{ \sum_{i=1}^T \sum_{j=1}^T | g(\bx_i; \btheta_0)^{\intercal} g(\bx_j; \btheta_0)/m  - \mathbf{M}_{i,j} |^2   } \\
&\leq T \epsilon.    
\end{aligned}
\]
Then, we have 
\[
\begin{aligned}
 &\log \det(\mathbf{I} + \mathbf{G} \mathbf{G}^{\intercal} /\lambda) \\
& = \log \deter ( \mathbf{I} + \mathbf{M} \lambda +  (\mathbf{G} \mathbf{G}^{\intercal} - \mathbf{M})/ \lambda  ) \\
& \leq   \log \deter ( \mathbf{I} +  \mathbf{M} \lambda) + \langle ( \mathbf{I} +  \mathbf{M} \lambda )^{-1},   (\mathbf{G} \mathbf{G}^{\intercal} - \mathbf{M})/ \lambda  \rangle \\
& \leq \log \deter(  \mathbf{I} +  \mathbf{M} \lambda) +  \| ( \mathbf{I} +  \mathbf{M} \lambda )^{-1}    \|_{F} \|  \mathbf{G} \mathbf{G}^{\intercal} - \mathbf{M}  \|_F / \lambda \\
& \leq  \log \deter(  \mathbf{I} +  \mathbf{M} \lambda) +  \sqrt{T}\|  \mathbf{G} \mathbf{G}^{\intercal} - \mathbf{M}  \|_F / \lambda \\
&\leq   \log \deter(  \mathbf{I} +  \mathbf{M} \lambda) + \lambda^{-1}\\
&= \widetilde{P} \log ( 1 + T/\lambda)+  \lambda^{-1}. 
\end{aligned}
\]
The first inequality is because the concavity of $\log \deter$ ; The third inequality is due to $  \| ( \mathbf{I} +  \mathbf{M} \lambda )^{-1} \|_{F} \leq  \| \mathbf{I}^{-1} \|_{F}  \leq \sqrt{T}$; The last inequality is because of the choice the $m$; The last equality is because  of the Definition \ref{def:dim}.
\newline

\begin{lemma} [ Lemma 4.1 in \cite{ntk2019generalization} ]  \label{lemma:f}
There exist constants $ \{ \bar{C}_{i=1}^3 \} \geq 0 $ such that for any $\delta \geq 0$, if $\tau$ satisfies that 
\[ 
\tau \leq \bar{C}_2 L^{-6}[\log m]^{-3/2},
\]
then with probability at least $1-\delta$, for all $\btheta^1, \btheta^2$ satisfying $\|\btheta^1- \btheta_0\| \leq \tau, \|\btheta^2 -\btheta_0 \| \leq \tau$ and for any $\bxt \in \{\bxt\}_{t=1}^T $, we have 
\[
| f(\bx; \btheta^1) - f(\bx; \btheta^2) - \langle ( g(\bx; \btheta^2 ), \btheta^1 -\btheta^2 )   \rangle  | \leq   \bar{C}_3  \tau^{4/3} L^3 \sqrt{m \log m}.  
\]
\end{lemma}

\begin{lemma}  [Lemma B.3 in \cite{ntk2019generalization} ] \label{lemma:g}
There exist constants $\{ C_i \}_{i=1}^2$ such that for any $\delta >0$, if $\tau$ satisfies that 
\[
 \tau \leq C_1 L^{-6} (\log m)^{-3/2},
\]
then, with probability at least $1-\delta$, for any $\| \btheta - \btheta_0\| \leq \tau$ and $\bxt \in \{\bx_t\}_{t=1}^T$ we have $\| g(\bx_t; \btheta) \|_2 \leq C_2 \sqrt{mL}$.
\end{lemma}

\begin{lemma} [Lemma B.2 in \cite{zhou2020neural} ] \label{lemma:b2}
For the $L$-layer full-connected network $f$, there exist constants $\{C_i\}_{i=1}^{5} \geq 0$ such that for $\delta >0$, if for all $t \in [T]$, $\eta, m$ satisfy
\[
\begin{aligned}
&2 \sqrt{t/(m\lambda)} \geq C_1 m^{-3/2}L^{-3/2}[\log (TL^2/\delta)]^{3/2}, \\ 
&2\sqrt{t/(m\lambda)} \leq C_2 \min \{ L^{-6} [ \log m]^{-3/2},  (m(\lambda \eta)^2 L^{-6} t^{-1} (\log m)^{-1} )^{3/8}\}, \\
&\eta \leq C_3 (m \lambda + tmL)^{-1},\\
&m^{1/6} \geq C_4 \sqrt{\log m} L^{7/2}t^{7/6} \lambda^{-7/6}(1+ \sqrt{t/\lambda}),
\end{aligned}
\]
then, with probability at least $1-\delta$, it has 
\[
\begin{aligned}
&\|\btheta_t - \btheta_0\| \leq 2 \sqrt{t/(m\lambda)}\\
&\|\btheta_t - \btheta_0 - \btheta'\| \leq  \\
&(1 - \eta m \lambda)^{J/2} \sqrt{t/(m\lambda)} + C_5 m^{-2/3} \sqrt{\log m}L^{7/2}t^{5/3} \lambda^{-5/3}(1+\sqrt{t/\lambda}).
\end{aligned}
\]
\end{lemma}

\begin{lemma} [Theorem 5 in \cite{allen2019convergence}] \label{lemma:b3}
With probability at least $1-\delta$, there exist constants $C_1, C_2$ such that if $\tau  \leq C_1 L^{-9/2} \log^{-3}m$, for $\|\btheta_t - \btheta_0\|_2 \leq \tau$, we have 
\[
 \|  g(\bx_t; \btheta_t) -  g(\bx_t; \btheta_0)\|_2  \leq C_2 \sqrt{\log m} \tau^{1/3} L^3 \|g (\bx_t; \btheta_0 )\|_2.    
\]

\end{lemma}

\subsection{Empirical Setting of UCB}
UCB is the key component of MuFasa, determining the empirical performance. $UCB(\mathbf{X}_t)$ (Lemma \ref{lemma:iucb}) can be formulated as
\[
UCB(\mathbf{X}_t) = (1 - \lambda) \|g(\bx; \btheta_t)\|_{\mathbf{A}_t^{-1}}  + \lambda  \|g(\bx; \btheta_0)\|_{\mathbf{A}_t^{'-1}}  
\]
where $g(\bx; \btheta_0)$ is the gradient at initialization, $g(\bx; \btheta_t)$ is the gradient after $k$ iterations of gradient descent, and $\lambda$ is a tunable parameter to trade off between them.
Intuitively, $g(\bx; \btheta_0)$ has more bias as the weights of neural network function $f$ are randomness initialized, which brings more exploration portion in decision making of each round. In contrast,  $g(\bx; \btheta_t)$ has more variance as the weights should be nearer to the optimum after gradient descent.

In the setting where the set of arms are fixed, given an arm $\bx_i$, let $m_i$ be the number of rounds that $\bx_i$ has been played before. When $m_i$ is small, the learner should explore $\bx_i$ more ($\lambda$ is expected to be large). Instead, when $m_i$ is large, the leaner does not need more exploration on it. Therefore, $\lambda$ can be defined as a decreasing function with respect to $m_i$, such as $\frac{1}{\sqrt{m_i+1}}$ and $\frac{1}{\log (m_i+1)}$. In the setting without this condition, we can set $\lambda$ as a decreasing function with respect to the number of rounds $t$, such as $\frac{1}{\sqrt{t+1}}$ and $\frac{1}{\log (t+1)}$.

\subsection{Future Work}

This paper introduces a novel bandit problem, multi-facet contextual bandits. Based on empirical experiments and theoretical analysis, at least three factors play important roles in multi-facet bandits: (1) As all bandits are serving the same user, mutual influence does exist among bandits. Thus, how to extract and harness the mutual influence among bandits is one future-work direction; (2) The weights of bandits usually determine the final reward, since each bandit has a different impact on the serving user. For example, in the diabetes case, the bandit for selecting medications should have a higher weight than the bandit for recommending diabetes articles; (3) Sub-rewards contain crucial information and thus exploiting sub-rewards matters in this problem.

To solve multi-facet bandit problem, we propose the algorithm, MuFasa. However, there are three limitations implicitly shown in this paper: (1) For leveraging the sub-rewards, we assume the $\bar{C}$-lipschitz continuity of final reward function, while $\bar{C}$ is usually unknown; (2) In the selection criteria of MuFasa, it needs to select one arm with the highest score from all possible combinations $\mathbf{S}_t$ (Eq. (2) and Eq. (5)). However, the size of $\mathbf{S}_t$ grows exponentially with respect to the number of bandits $K$. (3) MuFasa only uses fully-connected work to exploit rewards, while many more advanced neural network models exist such as CNN, residual net, self-attention, etc.

\end{document}